\documentclass{article}

\usepackage{microtype}
\usepackage{graphicx}
\usepackage{pgfplots}
\usepgfplotslibrary{groupplots}
\usepackage{subfigure}
\usepackage{booktabs} 
\usepackage{array}
\usepackage{amsmath}
\usepackage{multirow}
\usepackage{multicol}
\usepackage{xcolor}
\usepackage{threeparttable}
\usepackage[noend]{algorithm2e}
\usepackage{url}
 
\usepackage{hyperref}

\usepackage[noend]{algorithmic}

\usepackage[accepted]{icml2021}

\def\code#1{\texttt{{\footnotesize#1}}}
\newcommand{\autopipe}{\texttt{PipeTransformer}}

\icmltitlerunning{PipeTransformer: Automated Elastic Pipelining for Distributed Training of Transformers}

\begin{document}

\twocolumn[
\icmltitle{PipeTransformer: Automated Elastic Pipelining for \\ Distributed Training of Transformers}\vspace{-1em}



\icmlsetsymbol{equal}{*}

\begin{icmlauthorlist}
\icmlauthor{Chaoyang He}{usc}
\icmlauthor{Shen Li}{fb}
\icmlauthor{Mahdi Soltanolkotabi}{usc}
\icmlauthor{Salman Avestimehr}{usc}
\end{icmlauthorlist}\vspace{-1em}

\icmlaffiliation{usc}{University of Southern California}
\icmlaffiliation{fb}{Facebook AI}
\icmlcorrespondingauthor{Chaoyang He}{chaoyang.he@usc.edu}

\icmlkeywords{Machine Learning, ICML}

\vskip 0.3in
]



\printAffiliationsAndNotice{}  

\begin{abstract}
The size of Transformer models is growing at an unprecedented rate. It has taken less than one year to reach trillion-level parameters since the release of GPT-3 (175B). 
Training such models requires both substantial engineering efforts and enormous computing resources, which are luxuries most research teams cannot afford. 
In this paper, we propose \autopipe, which leverages automated elastic pipelining for efficient distributed training of Transformer models. 
In \autopipe, we design an adaptive on the fly freeze algorithm that can identify and freeze some layers gradually during training, and an elastic pipelining system that can dynamically allocate resources to train the remaining active layers.
More specifically, \autopipe\  automatically excludes frozen layers from the pipeline, packs active layers into fewer GPUs, and forks more replicas to increase data-parallel width. 
We evaluate \autopipe\ using Vision Transformer (ViT) on ImageNet and BERT on SQuAD and GLUE datasets. Our results show that compared to the state-of-the-art baseline, \autopipe\ attains up to $2.83$-fold speedup without losing accuracy.
We also provide various performance analyses for a more comprehensive understanding of our algorithmic and system-wise design. Finally, we have modularized our training system with flexible APIs and made the source code publicly available.

\end{abstract}

\section{Introduction}
\label{sec:intro}

Large Transformer models~\cite{gpt3, gshard} have powered accuracy breakthroughs in both natural language processing and computer vision. GPT-3 hit a new record high accuracy for nearly all NLP tasks. Vision Transformer (ViT) \cite{dosovitskiy2020image} also achieved 89\% top-1 accuracy in ImageNet, outperforming state-of-the-art convolutional networks ResNet-152 \cite{he2016deep} and EfficientNet \cite{tan2019efficientnet}. To tackle the growth in model sizes, researchers have proposed various distributed training techniques, including parameter servers~\cite{ps, byteps, parallax}, pipeline parallel~\cite{gpipe, hetpipe, pipedream}, intra-layer parallel~\cite{gshard, meshtf, megatron}, and zero redundancy data parallel~\cite{zero}. 
\vspace{-0.05em}
\begin{figure}[h!]
    \centering
    \includegraphics[width=1\linewidth]{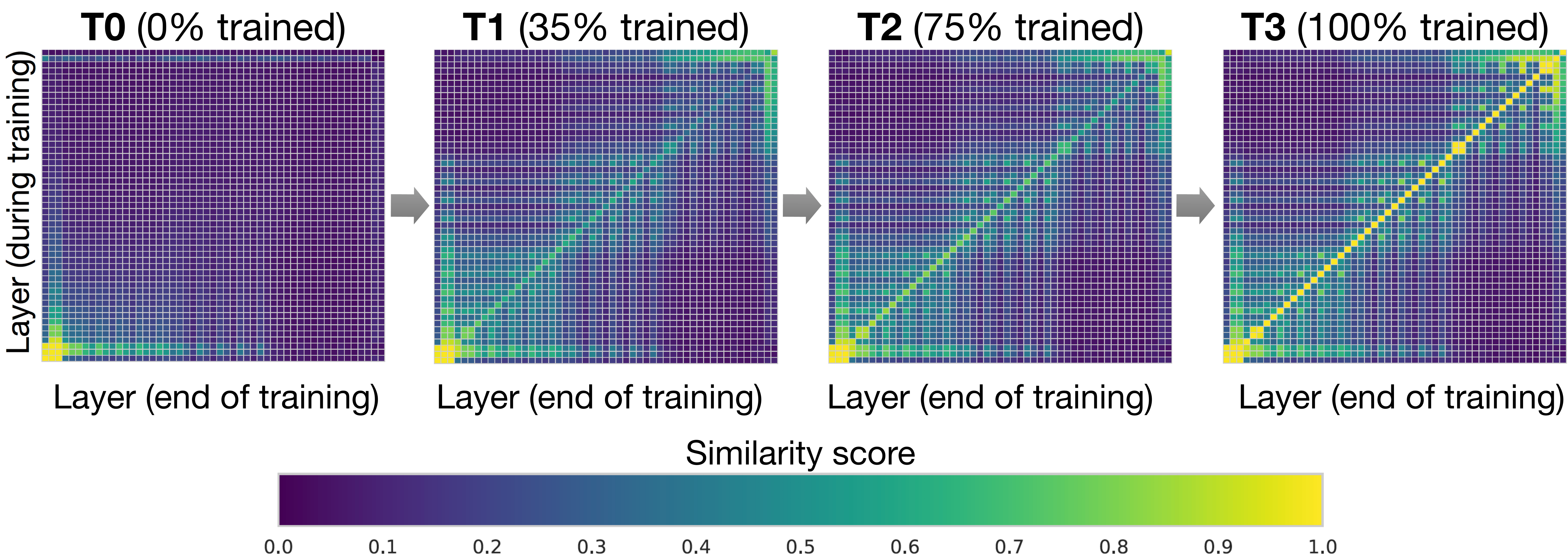}
    \vspace{-0.5cm}
    \caption{Interpretable Freeze Training: DNNs converge bottom up (Results on CIFAR10 using ResNet). Each pane shows layer-by-layer similarity using SVCCA \cite{Raghu2017SVCCASV}.}
    \label{fig:intepretable_freeze}
    \vspace{-0.4cm}
\end{figure}
\vspace{-0.2em}

\begin{figure*}[h!]
    \centering
    \makebox[\textwidth][c]{\includegraphics[width=\textwidth]{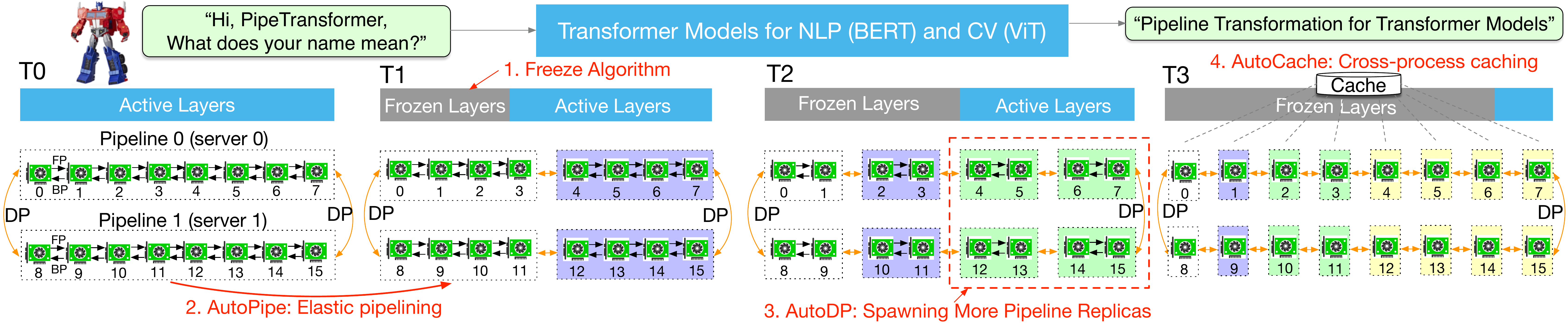}}
    \vspace{-0.8cm}
    \caption{The process of \autopipe's automated and elastic pipelining to accelerate distributed training of Transformer models}
    \vspace{-0.5cm}
    \label{fig:PipeTransformer}
\end{figure*}


Existing distributed training solutions, however, only study scenarios where all model weights are required to be optimized throughout the training (i.e., computation and communication overhead remains relatively static over different iterations). 
Recent works on \textit{freeze training} \cite{Raghu2017SVCCASV,NIPS2018_7815,reservoir} suggest that parameters in neural networks usually converge from the bottom-up (i.e., not all layers need to be trained all the way through training). Figure~\ref{fig:intepretable_freeze} shows an example of how weights gradually stabilize during training in this approach. 
This observation motivates us to utilize freeze training for distributed training of Transformer models to accelerate training by dynamically allocating resources to focus on a shrinking set of active layers.
Such a layer freezing strategy is especially pertinent to pipeline parallelism, as excluding consecutive bottom layers from the pipeline can reduce computation, memory, and communication overhead.


In this paper, we propose \code{PipeTransformer}, an elastic pipelining training acceleration framework that automatically reacts to frozen layers by dynamically transforming the scope of the pipelined model and the number of pipeline replicas. To the best of our knowledge, this is the first paper that studies layer freezing in the context of both pipeline and data-parallel training. Figure \ref{fig:PipeTransformer} demonstrates the benefits of such a combination. First, by excluding frozen layers from the pipeline, the same model can be packed into fewer GPUs, leading to both fewer cross-GPU communications and smaller pipeline bubbles. Second, after packing the model into fewer GPUs, the same cluster can accommodate more pipeline replicas, increasing the width of data parallelism. More importantly, the speedups acquired from these two benefits are multiplicative rather than additive, further accelerating the training.

The design of \code{PipeTransformer} faces four major challenges.
 First, the freeze algorithm must make \textit{on the fly} and adaptive freezing decisions; however, existing work~\cite{Raghu2017SVCCASV} only provides a posterior analysis tool.
 Second, the efficiency of pipeline re-partitioning results is influenced by multiple factors, including partition granularity, cross-partition activation size, and the chunking (the number of micro-batches) in mini-batches, which require reasoning and searching in a large solution space. 
 Third, to dynamically introduce additional pipeline replicas, \code{PipeTransformer} must overcome the static nature of collective communications and avoid potentially complex cross-process messaging protocols when onboarding new processes (one pipeline is handled by one process).
 Finally, caching can save time for repeated forward propagation of frozen layers, but it must be shared between existing pipelines and newly added ones, as the system cannot afford to create and warm up a dedicated cache for each replica.

\code{PipeTransformer} is designed with four core building blocks to address the aforementioned challenges.
 First, we design a tunable and adaptive algorithm to generate signals that guide the selection of layers to freeze over different iterations (Section \ref{sec:freeze}). Once triggered by these signals, our elastic pipelining module \code{AutoPipe}, then packs the remaining active layers into fewer GPUs by taking both activation sizes and variances of workloads across heterogeneous partitions (frozen layers and active layers) into account. It then splits a mini-batch into an optimal number of micro-batches based on prior profiling results for different pipeline lengths (Section \ref{sec:auto_pipe}). Our next module, \code{AutoDP}, spawns additional pipeline replicas to occupy freed-up GPUs and maintains hierarchical communication process groups to attain dynamic membership for collective communications (Section \ref{sec:auto_dp}). Our final module, \code{AutoCache}, efficiently shares activations across existing and new data-parallel processes and automatically replaces stale caches during transitions (Section \ref{sec:auto_cache}). 
 
Overall, \autopipe\ combines the \code{Freeze Algorithm}, \code{AutoPipe}, \code{AutoDP} and \code{AutoCache} modules to  provide a significant training speedup.
We evaluate \autopipe\  using Vision Transformer (ViT) on ImageNet and BERT on GLUE and SQuAD datasets. Our results show that \autopipe\ attains up to $2.83$-fold speedup without losing accuracy. We also provide various performance analyses for a more comprehensive understanding of our algorithmic and system-wise design. 
Finally, we have also developed open-source flexible APIs for \autopipe\, which offer a clean separation among the freeze algorithm, model definitions, and training accelerations, allowing for transferability to other algorithms that require similar freezing strategies. The source code is made publicly available.

\section{Overview}
\label{sec:method}


\subsection{Background and Problem Setting}
\label{sec:bg_problem}
Suppose we aim to train a massive model in a distributed training system where the \textit{hybrid of pipelined model parallelism and data parallelism} is used to target scenarios where either the memory of a single GPU device cannot hold the model, or if loaded, the batch size is small enough to avoid running out of memory. More specifically, we define our settings as follows:


\textbf{Training task and model definition.} We train Transformer models (e.g., Vision Transformer \cite{dosovitskiy2020image}, BERT \cite{devlin2018bert}) on large-scale image or text datasets. The Transformer model $\mathcal{F}$ has $L$ layers, in which the $i$th layer is composed of a forward computation function $f_i$ and a corresponding set of parameters, $\mathbf{w}_i$. With this definition, the overall model is  $\mathcal{F}=f_{0}(\mathbf{w}_0) \circ \ldots \circ f_{L-1}(\mathbf{w}_{L-1})$.
The model size is $S$, and the batch size is set to $N_{bs}$. 

\textbf{Training infrastructure.} Assume the training infrastructure contains a GPU cluster that has $N$ GPU servers (i.e. nodes). Each node has $I$ GPUs. Our cluster is homogeneous, meaning that each GPU and server have the same hardware configuration. Each GPU's memory capacity is $M_\text{GPU}$. Servers are connected by a high bandwidth network interface such as \texttt{InfiniBand} interconnect.


\textbf{Pipeline parallelism.} In each machine, we load a model $\mathcal{F}$ into a pipeline $\mathcal{P}$ which has $K$ partitions ($K$ also represents the pipeline length). The $k$th partition $p_k$ consists of consecutive layers $p_k=f_{i}(\mathbf{w}_i) \circ \ldots \circ f_{j}(\mathbf{w}_{j})$, and $\mathcal{P}=p_{0} \circ \ldots \circ p_{K-1}$. We assume each partition is handled by a single GPU device. $1 \leq K \leq I$, meaning that we can build multiple pipelines for multiple model replicas in a single machine. We assume all GPU devices in a pipeline belong to the same machine. Our pipeline is a synchronous pipeline, which does not involve stale gradients, and the number of micro-batches is $M$. In the Linux OS, each pipeline is handled by a single process. We refer the reader to \code{GPipe} \cite{gpipe} for more details.

\textbf{Data parallelism.} $\code{DDP}$ \cite{ddp} is a cross-machine distributed data parallel process group within $R$ parallel workers. Each worker is a pipeline replica (a single process). The $r$th worker's index (ID) is rank $r$. For any two pipelines $\mathcal{P}^{(r_i)}$ and $\mathcal{P}^{(r_j)}$ in $\code{DDP}$, $r_i$ and $r_j$ can belong to either the same GPU server or different GPU servers, and they can exchange gradients with the \code{AllReduce} algorithm.

Under these settings, our goal is to accelerate training by leveraging \emph{freeze training}, which does not require all layers to be trained throughout the duration of the training. Additionally, it may help save computation, communication, memory cost, and potentially prevent overfitting by consecutively freezing layers. However, these benefits can only be achieved by overcoming the four challenges of designing an adaptive freezing algorithm, dynamical pipeline re-partitioning, efficient resource reallocation, and cross-process caching, as discussed in the introduction.
We next describe our overall design, named \autopipe, which can address these challenges. 



\subsection{Overall Design}
\label{sec:overall_design}

\begin{figure}[h!]
    \vspace{-0.2cm}
    \centering
    \includegraphics[width=1.0\linewidth]{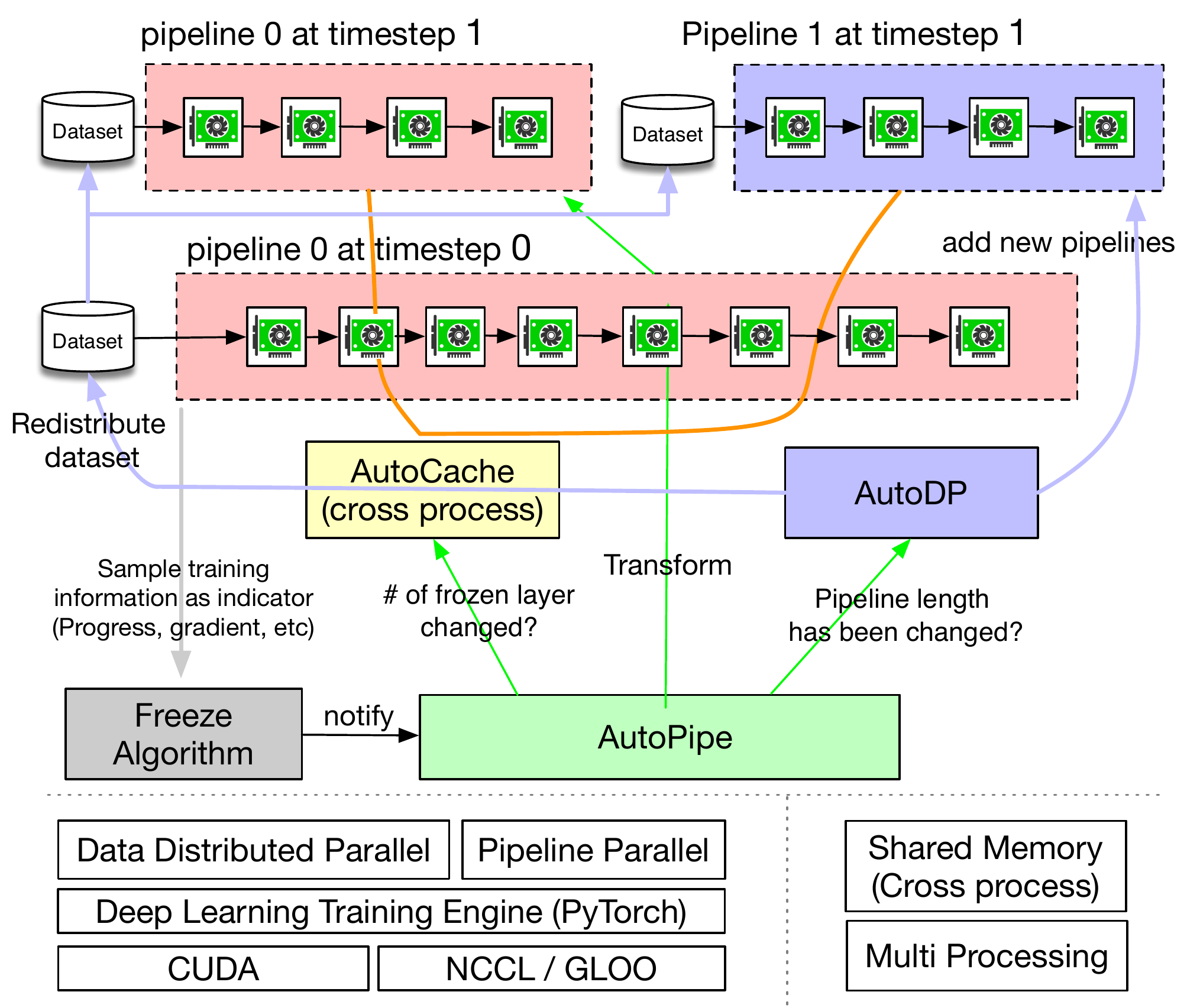}
    \vspace{-0.5cm}
    \caption{Overview of \autopipe\ Training System}
    \label{fig:overview_design}
    \vspace{-0.3cm}
\end{figure}

\autopipe\ co-designs an on the fly freeze algorithm and an automated elastic pipelining training system that can dynamically transform the scope of the pipelined model and the number of pipeline replicas.
The overall system architecture is illustrated in Figure \ref{fig:overview_design}. To support \autopipe's elastic pipelining, we maintain a customized version of \code{PyTorch Pipe}~\cite{kim2020torchgpipe}. 
For data parallelism, we use \code{PyTorch DDP} \cite{ddp} as a baseline. Other libraries are standard mechanisms of an operating system (e.g., \code{multi-processing}) and thus avoid specialized software or hardware customization requirements. To ensure the \textit{generality} of our framework, we have decoupled the training system into four core components: \texttt{freeze algorithm}, \texttt{AutoPipe}, \texttt{AutoDP}, and \texttt{AutoCache}. The freeze algorithm (grey) samples indicators from the training loop and makes layer-wise freezing decisions, which will be shared with \texttt{AutoPipe} (green). \texttt{AutoPipe} is an elastic pipeline module that speeds up training by excluding frozen layers from the pipeline and packing the active layers into fewer GPUs (pink), leading to both fewer cross-GPU communications and smaller pipeline bubbles. Subsequently, \texttt{AutoPipe} passes \textit{pipeline length} information to \texttt{AutoDP} (purple), which then spawns more pipeline replicas to increase data-parallel width, if possible. The illustration also includes an example in which AutoDP introduces a new replica (purple). \texttt{AutoCache} (orange edges) is a cross-pipeline caching module, as illustrated by connections between pipelines. The source code architecture is aligned with Figure \ref{fig:overview_design} for readability and generality.

\section{Algorithm and System Design}
This section elaborates on the four main algorithmic and system-wise design components of \autopipe.

\subsection{Freeze Algorithm}
\label{sec:freeze}



\begin{figure}[h!]
    \vspace{-0.2cm}
    \centering
    \includegraphics[width=0.75\linewidth]{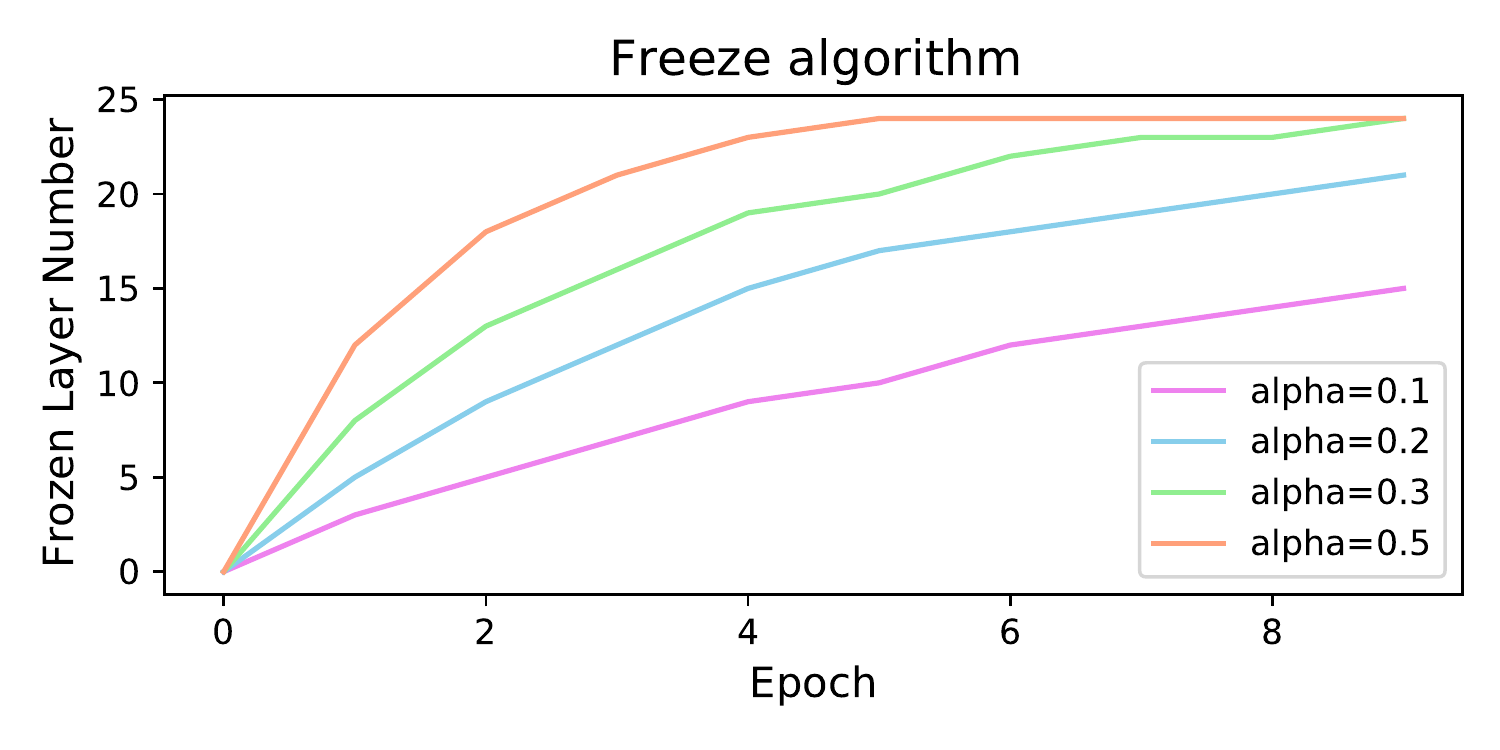}
    \vspace{-15pt}
    \caption{Freeze Algorithm Using Different $\alpha$ }
    \label{fig:freeze}
    \vspace{-0.3cm}
\end{figure}

The freeze algorithm must be lightweight and able to make decisions on the fly. This excludes existing layer-wise training approaches such as SVCCA~\cite{Raghu2017SVCCASV} which require full training states and heavy posterior measurements. We propose an adaptive on the fly freeze algorithm to define $L_{\text{frozen}}^{(T)}$ at timestep $T$ as follows:
\vspace{-0.2em}
\begin{equation}
\begin{split}
\footnotesize
\label{eq:freeze}
\text{$\min \Bigg( L_{\text{frozen}}^{(T-1)} + \alpha(L - L_{\text{frozen}}^{(T-1)}), \operatornamewithlimits{argmin}\limits_{\ell \in \{L_{\text{frozen}}^{(T-1)}, ..., L\}} \left\|\boldsymbol{g}_{\ell}^{(T)}\right\| \Bigg)$} \\
\text{where $T \geq 1$, $L_{\text{frozen}}^{(0)}=0$, and $\alpha \in (0,1)$}
\end{split}
\end{equation}
\vspace{-1.0em}

where $g_{\ell}^{(T)}$ is the gradient for layer $\ell$ at iteration $T$, and $\left\|\boldsymbol{g}_{\ell}^{(T)}\right\|$ is its norm. The intuition behind the second term in the $\min$ function is that the layer with the smallest gradient norm converges first. To stabilize training, we enforce an upper bound $L_{\text{frozen}}^{(T-1)} + \alpha(L - L_{\text{frozen}}^{(T-1)})$ for the number of frozen layers, which is a geometric sequence containing a hyper-parameter $\alpha$. This essentially freezes an $\alpha$ fraction of the remaining active layers. To illustrate the impact of $\alpha$, we rewrite the equation as: $L_{\text{frozen}}^{(T)} = (1 - \alpha)^{T}[\frac{{\alpha}L}{1-\alpha} + \sum_{t=2}^{T}{\frac{{\alpha}L}{(1-\alpha)^t}}]$ (see Appendix 
for the derivation), and draw the curve of this function in Figure \ref{fig:freeze}. As we can see, a larger $\alpha$ leads to a more aggressive layer freezing. Therefore, Equation \ref{eq:freeze} calculates the number of frozen layers at timestep $T$ using both the gradient norm and a tunable argument $\alpha$.

The $\alpha$ parameter controls the trade-off between accuracy and training speed. This algorithm is also analogous to learning rate (LR) decay. Both algorithms use a scheduler function during training, and take the progress of training as an indicator. 
The difference is that the above freeze algorithm also takes gradient norm into account, making the algorithm simple and effective. Other freezing strategies can be easily plugged into the our training system. Indeed, we plan to investigate other strategies in our future work.  
\subsection{AutoPipe: Elastic Pipelining}
\label{sec:auto_pipe}

\begin{algorithm*}[h!]
   \caption{\texttt{AutoPipe} Algorithm}
   \label{alg:transformation algorithm}
\begin{multicols}{2}   
\footnotesize
\begin{algorithmic}[1]
   \STATE {\bfseries Input:} model $\mathcal{F}$, layer number $L$ and $L_{\text{frozen}}$, pipeline length $K$, frozen layer cost factor $\lambda_{\text{frozen}}$
   \STATE {\bfseries Return:}
   model $\mathcal{F}_{\text{frozen}}$, model $\mathcal{F}_{\text{pipe}}$, updated $K$;

   \STATE \code{\colorbox[gray]{0.85}{def m\_partition($\mathcal{F}$,$L$, $L_{\text{frozen}}$):}} \textit{//see \ref{sec:model_partition}}
   \STATE $\mathcal{F}_{\text{frozen}}=\code{Sequential()}$; model size $S_\text{frozen} = 0$
   \STATE $\mathcal{F}_{\text{pipe}}=\code{Sequential()}$; per-layer  size $S_\text{pipe} = \code{[]}$
   \FOR{layer index = $L_{\text{frozen}}$ {\bfseries to} $L$}
   \STATE \colorbox{green!20}{${f_{\text{ATT}}}_i, {f_{\text{MLP}}}_i \leftarrow f_i $}
   \STATE $\mathcal{F}_{\text{pipe}}.\code{append}({f_{\text{ATT}}}_i); S_\text{pipe}.\code{append}(\code{m\_size}({f_{\text{ATT}}}_i))$
   \STATE $\mathcal{F}_{\text{pipe}}.\code{append}({f_{\text{MLP}}}_i); S_\text{pipe}.\code{append}(\code{m\_size}({f_{\text{MLP}}}_i))$
   \ENDFOR
   \STATE {\bfseries return} $\mathcal{F}_{\text{frozen}}$,$S_\text{frozen}$,$\mathcal{F}_{\text{pipe}}$,$S_\text{pipe}$

   \STATE \colorbox[gray]{0.85}{\code{def load\_balance}($\mathcal{F}_{\text{pipe}}$, $S_\text{pipe}$, $K$):} \textit{//Section \ref{sec:model_partition}}
   \STATE $B_{L}$=\code{dict}(), $B_{S}$=\code{dict}() \textit{// balanced L and S}
   \STATE $L_{\text{assigned}} = 0$; $S_{\text{total}}$ = \code{sum}($S_\text{pipe}$)
   \FOR{partition index = $k$ {\bfseries to} $K$}
   \STATE \code{mean}=$S_{\text{total}}$/($K$ - $k$); 
   \STATE \code{var=np.var}($S_\text{pipe}$[$L_{\text{assigned}}$:])/($K$ - $k$)
      \FOR{sublayer index i = $L_{\text{assigned}}$ {\bfseries to} \code{len}($S_\text{pipe}$)}
          \STATE $S_k$ = $S_\text{pipe}$[i]
          \STATE \code{criterion}=$B_{S}$[i]-$S_\text{frozen}$(1.0-\colorbox{red!18}{$\lambda_{\text{frozen}}$})+$S_k$
          \IF{\code{criterion < mean + var}}
              \STATE $B_{S}$+=$S_k$; $B_{L}$+=1; $L_{\text{assigned}}$+=1; $S_{\text{total}}$-=$S_k$
          \ELSE
            \STATE \code{break}
          \ENDIF
      \ENDFOR
   \ENDFOR
   \STATE {\bfseries return} $B_{L}$, $B_{S}$
   
   \STATE $\mathcal{F}_{\text{frozen}}$,$S_\text{frozen}$,$\mathcal{F}_{\text{pipe}}$,$S_\text{pipe}$ = \code{m\_partition}($\mathcal{F}$,$L$, $L_{\text{frozen}}$)
   \WHILE{$K \geq 2$}
   \STATE $B_{L}$, $B_{S}$ = \code{load\_balance}($\mathcal{F}_{\text{pipe}}$, $S_\text{pipe}$, $K/2$)
   \STATE $B_{S}$[0] -= $S_\text{frozen}$(1.0 - $\lambda_{\text{frozen}}$); 
   \STATE $M_{GPU}^{(T)}$ = \code{max}($B_{S}$) \textit{ //Equation \ref{eq:compression}}
  \IF{\colorbox{blue!18}{$M_{GPU}^{(T)} < M_{GPU}^{(0)}$}}
        \STATE \code{$K$=$K$/2}
  \ELSE
        \STATE break
  \ENDIF
  \ENDWHILE
  \STATE load $\mathcal{F}_{\text{frozen}}$ and $\mathcal{F}_{\text{pipe}}$ to $K$ GPUs using $B_{S}$ and $B_{L}$
  \STATE \code{Pipe($\mathcal{F}_{\text{pipe}}$, chunks=\colorbox{yellow!20}{\code{get\_optimal\_chunks}}($K$))}
\end{algorithmic}
\end{multicols}
\label{alg:autopipe}
\vspace{-0.2cm}
\end{algorithm*}

Triggered by the freeze algorithm, \code{AutoPipe} can accelerate training by excluding frozen layers from the pipeline and packing the active layers into fewer GPUs. This section elaborates on the key components of \code{AutoPipe} that dynamically partition pipelines, minimize the number of pipeline devices and optimize mini-batch chunk size accordingly. Algorithm \ref{alg:transformation algorithm} presents the pseudo-code. 

\subsubsection{Balanced Pipeline Partitioning}
\label{sec:model_partition}
\begin{figure}[!h]
    \centering
    \includegraphics[width=1\linewidth]{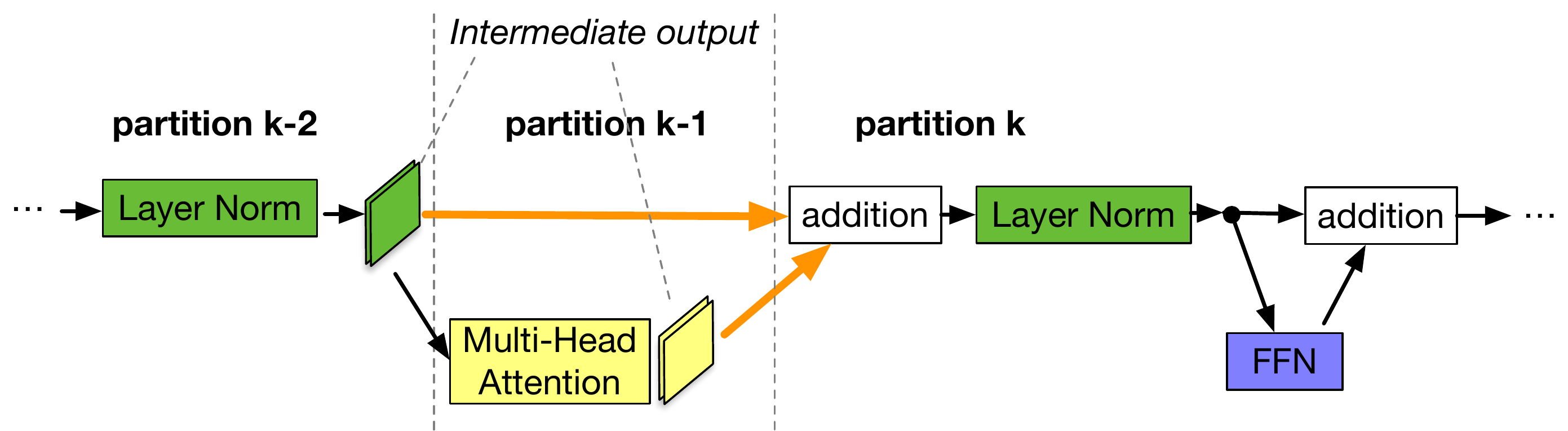}
    \vspace{-0.5cm}
    \caption{Partition boundary is in the middle of a skip connection}
    \label{fig:skip_connection}
    \vspace{-0.3cm}
\end{figure}
Balancing computation time across partitions is critical to pipeline training speed, as skewed workload distributions across stages can lead to stragglers, forcing devices with lighter workloads to wait (demonstrated by Section \ref{sec:speedup_breakdown}). However, maintaining optimally balanced partitions does not guarantee the fastest training speed because other factors also play a crucial role: 

1. \textit{Cross-partition communication overhead.}
Placing a partition boundary in the middle of a skip connection leads to additional communications since tensors in the skip connection must now be copied to a different GPU. For example, with BERT partitions in figure \ref{fig:skip_connection}, partition $k$ must take intermediate outputs from both partition $k-2$ and partition $k-1$. In contrast, if the boundary is placed after the \code{addition} layer, the communication overhead between partition $k-1$ and $k$ is visibly smaller.
Our measurements show that having cross-device communication is more expensive than having slightly imbalanced partitions (see the Appendix). Therefore, we do not consider breaking skip connections (highlighted separately as an entire attention layer ${f_{\text{ATT}}}_i$ and MLP layer ${f_{\text{MLP}}}_i$ in green at line 7 in Algorithm \ref{alg:autopipe}).

2. \textit{Frozen layer memory footprint.} During training, \code{AutoPipe} must recompute partition boundaries several times to balance two distinct types of layers: frozen layers and active layers. The frozen layer's memory cost is a fraction of that in active layers, given that the frozen layer does not need backward activation maps, optimizer states, and gradients. Instead of launching intrusive profilers to obtain thorough metrics on memory and computational cost, we define a tunable cost factor $\lambda_{\text{frozen}}$ to estimate the memory footprint ratio of a frozen layer over the same active layer. Based on empirical measurements in our experimental hardware, we set $\lambda_{\text{frozen}}$ to $\frac{1}{6}$.

Based on the above two considerations, \code{AutoPipe} balances pipeline partitions based on parameter sizes. More specifically, \code{AutoPipe} uses a greedy algorithm to allocate all frozen and active layers to evenly distribute partitioned sublayers into $K$ GPU devices. Pseudo code is described as the \code{load\_balance()} function in Algorithm \ref{alg:autopipe}. The frozen layers are extracted from the original model and kept in a separate model instance $\mathcal{F}_{\text{frozen}}$ in the first device of a pipeline.
Note that the partition algorithm employed in this paper is not the only option; \code{PipeTransformer} is modularized to work with any alternatives.

\subsubsection{Pipeline Compression}
\label{sec:pipe_compression}

Pipeline compression helps to free up GPUs to accommodate more pipeline replicas and reduce the number of cross-device communications between partitions. 
To determine the timing of compression, we can estimate the memory cost of the largest partition after compression, and then compare it with that of the largest partition of a pipeline at timestep $T=0$. To avoid extensive memory profiling, the compression algorithm uses the parameter size as a proxy for the training memory footprint. Based on this simplification, the criterion of pipeline compression is as follows:
\vspace{-0.5em}
\begin{equation}
\label{eq:compression}
\begin{split}
\text{compress the pipeline if } M_{GPU}^{(T)} \leq M_{GPU}^{(0)} \\
\text{where } M_{GPU}^{{(T)}} \Leftrightarrow \max _{k \in \{0, \cdots, K-1\} } S_{p_k}
\end{split}
\end{equation}
Once the freeze notification is received, \code{AutoPipe} will always attempt to divide the pipeline length $K$ by 2 (e.g., from 8 to 4, then 2). By using $\frac{K}{2}$ as the input, the compression algorithm can verify if the result satisfies the criterion in Equation (1). Pseudo code is shown in lines 25-33 in Algorithm \ref{alg:autopipe}. Note that this compression makes the acceleration ratio \textit{exponentially} increase during training, meaning that if a GPU server has a larger number of GPUs (e.g., more than 8), the acceleration ratio will be further amplified.

\vspace{-0.5em}
\begin{figure}[h!]
    \centering
    \includegraphics[width=\linewidth]{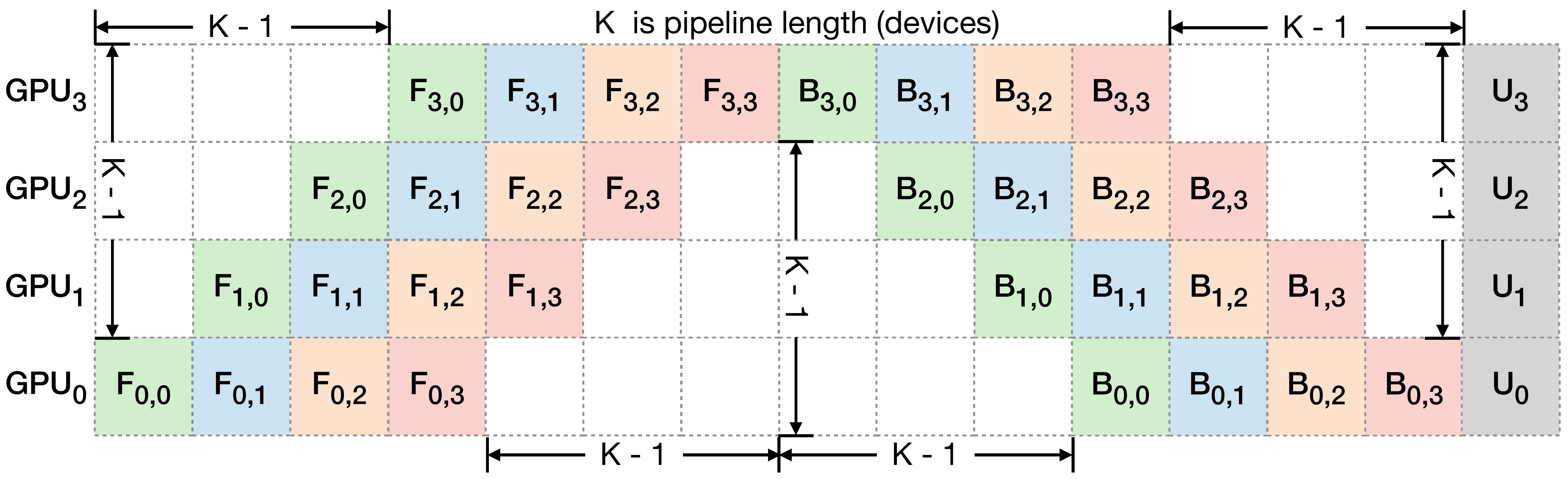}
    \vspace{-2em}
    \caption{Pipeline Bubble: $F_{d,b}$, $B_{d, b}$, and $U_d$ denote forward, backward, and the optimizer update of micro-batch $b$ on device $d$, respectively. The total bubble size in each iteration is $(K-1)$ times per micro-batch forward and backward cost.}
    \label{fig:bubble}
\end{figure}
\vspace{-1em}

Additionally, such a technique can also speed up training by shrinking the size of pipeline bubbles. To explain bubble sizes in a pipeline, Figure~\ref{fig:bubble} depicts how 4 micro-batches run through a 4-device pipeline ($K = 4$). In general, the total bubble size is $(K-1)$ times per micro-batch forward and backward cost (for further explanation, please refer to Appendix.

Therefore, it is clear that shorter pipelines have smaller bubble sizes. 

\subsubsection{Dynamic Number of Micro-batches}
\label{sec:micro_batches}

Prior pipeline parallel systems use a fixed number of micro-batches per mini-batch ($M$). \code{GPipe} suggests $M \geq 4 \times K$, where $K$ is the number of partitions (pipeline length). However, given that that \code{PipeTransformer} dynamically configures $K$, we find it to be sub-optimal to maintain a static $M$ during training. Moreover, when integrated with \code{DDP}, the value of $M$ also has an impact on the efficiency of \code{DDP} gradient synchronizations. Since \code{DDP} must wait for the last micro-batch to finish its backward computation on a parameter before launching its gradient synchronization, finer micro-batches lead to a smaller overlap between computation and communication (see Appendix for illustration). Hence, instead of using a static value, \code{PipeTransformer} searches for optimal $M$ on the fly in the hybrid of \code{DDP} environment by enumerating $M$ values ranging from $K$ to $6K$. For a specific training environment, the profiling needs only to be done once (see Algorithm~\ref{alg:autopipe} line 35). Section~\ref{sec:experiments} will provide performance analyses of $M$ selections.



\subsection{AutoDP: Spawning More Pipeline Replicas}
\label{sec:auto_dp}
As \code{AutoPipe} compresses the same pipeline into fewer GPUs, \code{AutoDP} can automatically spawn new pipeline replicas to increase data-parallel width. 

Despite the conceptual simplicity, subtle dependencies on communications and states require careful design. The challenges are threefold: 1. \code{DDP} \textit{Communication}: Collective communications in PyTorch \code{DDP} requires static membership, which prevents new pipelines from connecting with existing ones; 2. \textit{State Synchronization}: newly activated processes must be consistent with existing pipelines in the training progress (e.g., epoch number and learning rate), weights and optimizer states, the boundary of frozen layers, and pipeline GPU range; 3. \textit{Dataset Redistribution}: the dataset should be re-balanced to match a dynamic number of pipelines. This not only avoids stragglers but also ensures that gradients from all DDP processes are equally weighted.

\begin{figure}[h!]
    \vspace{-0.4cm}
    \centering
    \includegraphics[width = 1 \linewidth]{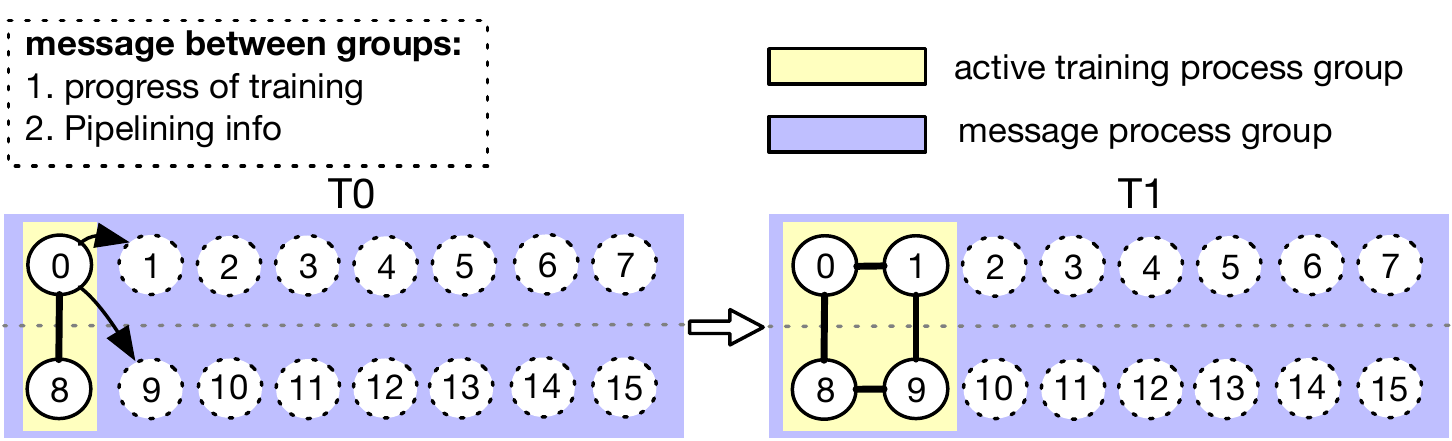}
    \vspace{-0.8cm}
    \caption{AutoDP: handling dynamical data parallel with messaging between double process groups (Process 0-7 belong to machine 0, while process 8-15 belong to machine 1)}
    \label{fig:autodp}
    \vspace{-0.33cm}
\end{figure}

To tackle these challenges, we create double communication process groups for \code{DDP}. As in the example shown in Figure \ref{fig:autodp}, the message process group (purple) is responsible for light-weight control messages and covers all processes, while the active training process group (yellow) only contains active processes and serves as a vehicle for heavy-weight tensor communications during training. The message group remains static, whereas the training group is dismantled and reconstructed to match active processes. 
In T0, only process \code{0} and \code{8} are active. During the transition to T1, process \code{0} activates processes \code{1} and \code{9} (newly added pipeline replicas) and synchronizes necessary information mentioned above using the message group. The four active processes then form a new training group, allowing static collective communications adaptive to dynamic memberships.
To redistribute the dataset, we implement a variant of \code{DistributedSampler} that can seamlessly adjust data samples to match the number of active pipeline replicas.

The above design also naturally helps to reduce \code{DDP} communication overhead. More specifically, when transitioning from T0 to T1, processes \code{0} and \code{1} destroy the existing \code{DDP} instances, and active processes construct a new \code{DDP} training group using $\mathcal{F}_{\text{pipe}}$ (\code{AutoPipe} stores $\mathcal{F}_{\text{frozen}}$ and $\mathcal{F}_{\text{pipe}}$ separately, introduced in Section \ref{sec:model_partition}). Discussion of communication cost can be found in Appendix.


\subsection{AutoCache: Cross-pipeline Caching}
\label{sec:auto_cache}

Caching activation maps from frozen layers can help further speed up training. This idea appears to be straightforward, but several caveats must be carefully addressed. 

\begin{figure}[h!]
    \centering
    \includegraphics[width = 0.75 \linewidth]{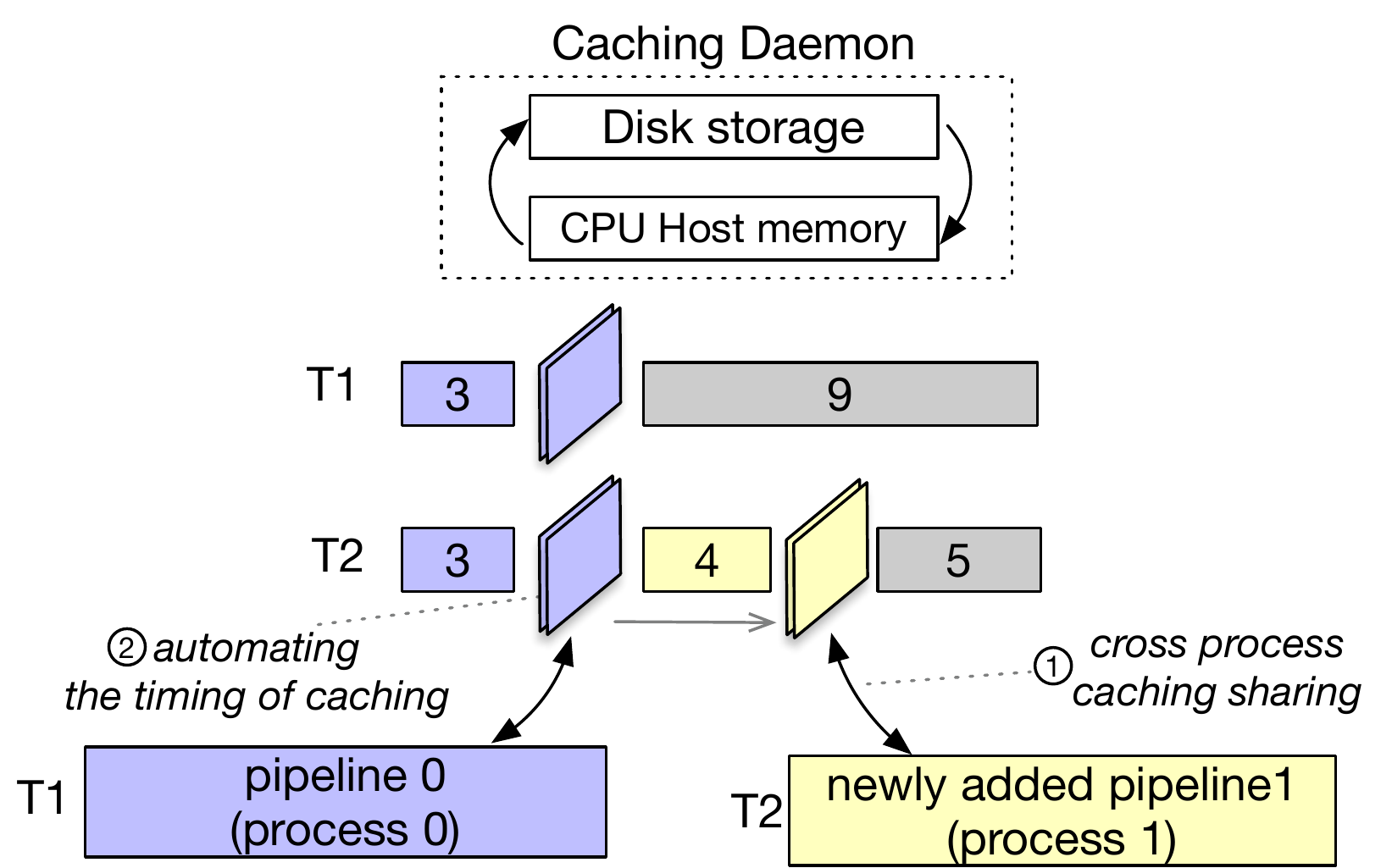}
    \vspace{-0.2cm}
    \caption{AutoCache}
    \label{fig:autocache}
    \vspace{-0.6cm}
\end{figure}

 \textbf{Cross-process caching.} The cache must be shared across processes in real time, as creating and warming up a dedicated cache for each model replica slow down the training. This is achieved by spawning a dedicated daemon process to hold cache in shared memory that all training processes can access in real time. Figure~\ref{fig:autocache} shows an example of the transition from T1 to T2, assuming T1 freezes 3 layers, T2 freezes 4 layers, and 5 layers remain active in T2. Immediately after the transition by \code{AutoDP}, the cache still holds cached activations from layer 3, which must be replaced by activations from layer 7. Therefore, all processes read their corresponding activations from the cache, feed them to the next 4 layers to compute activations for layer 7, then replace the existing cache with new activations for their samples accordingly. In this way, \code{AutoCache} can gradually update cached activations without running any sample through any frozen layers twice. 

 When the activations are too large to reside on CPU memory, \code{AutoCache} will also swap them to the disk and perform pre-fetching automatically. More details on the cross-process cache design can be found in the Appendix.

 \textbf{Timing of cache} is also important, as the cache can be slower than running the real forward propagation, especially if frozen layers are few and activations are large. To ensure that our training system can adapt to different hardware, model architecture, and batch size settings, \texttt{AutoCache} also contains a profiler that helps evaluate the appropriate transition to enable caching, and it only employs cached activations when the profiler suggests caching can speed up the forward pass. Performance analysis is provided at Section \ref{sec:timing_of_caching}.

\section{Experiments}
\label{sec:experiments}
This section first summarizes experiment setups and then evaluates \code{PipeTransformer} using computer vision and natural language processing tasks. More comprehensive results can be found in the Appendix.

\vspace{-0.2cm}
\subsection{Setup}

\paragraph{Hardware.} Experiments were conducted on 2 identical machines connected by InfiniBand CX353A ($5$GB/s), where each machine is equipped with 8 NVIDIA Quadro RTX 5000 (16GB GPU memory). GPU-to-GPU bandwidth within a machine (PCI 3.0, 16 lanes) is $15.754$GB/s.

\vspace{-0.3cm}
\paragraph{Implementation.} We used \code{PyTorch Pipe} as a building block, which has not yet been officially released at the time of writing of this paper. Hence, we used the developer version \code{1.8.0.dev20201219}. The BERT model definition, configuration, and related tokenizer are from \code{HuggingFace 3.5.0}. We implemented Vision Transformer using PyTorch by following its TensorFlow implementation. More details can be found in our source code.

\vspace{-0.3cm}
\paragraph{Models and Datasets.} Experiments employ two representative Transformers in CV and NLP: Vision Transformer (ViT) and BERT. ViT was run on an image classification task, initialized with pre-trained weights on ImageNet21K and fine-tuned on ImageNet and CIFAR-100. BERT was run on two tasks, text classification on the SST-2 dataset from the General Language Understanding Evaluation (GLUE) benchmark, and question answering on the SQuAD v1.1 Dataset (Stanford Question Answering) which is a collection of 100k crowdsourced question/answer pairs.

\vspace{-0.3cm}
\paragraph{Training Schemes.} Given that large models normally would require thousands of GPU-days (\emph{e.g.}, GPT-3) if trained from scratch, fine-tuning downstream tasks using pre-trained models has become a trend in CV and NLP communities. Moreover, \autopipe\ is a complex training system that involves multiple core components. Thus, for the first version of \autopipe\ system development and algorithmic research, it is not cost-efficient to develop and evaluate from scratch using large-scale pretraining. Therefore, experiments presented in this section focuses on pre-trained models. Note that since the model architectures in pre-training and fine-tuning are the same, \code{PipeTransformer} can serve both. We discussed pre-training results in the Appendix.

\vspace{-0.3cm}
\paragraph{Baseline.} Experiments in this section compares \code{PipeTransformer} to the state-of-the-art framework, a hybrid scheme of \code{PyTorch Pipe} (PyTorch’s implementation of GPipe~\cite{gpipe}) and \code{PyTorch DDP}. Since this is the first paper that studies accelerating distributed training by freezing layers, there are no perfectly aligned counterpart solutions yet.

\vspace{-0.3cm}
\paragraph{Hyper-parameters.} Experiments use ViT-B/16 (12 transformer layers, $16 \times 16$ input patch size) for ImageNet and CIFAR-100, BERT-large-uncased (24 layers) for SQuAD 1.1, and BERT-base-uncased (12 layers) for SST-2. With \code{PipeTransformer}, ViT and BERT training can set the per-pipeline batch size to around 400 and 64 respectively. Other hyperparameters (e.g., epoch, learning rate) for all experiments are presented in Appendix.

\subsection{Overall Training Acceleration}
We summarize the overall experimental results in Table \ref{table:speedup_cv}. Note that the speedup we report is based on a conservative $\alpha$ ($\frac{1}{3}$) value that can obtain comparable or even higher accuracy. A more aggressive $\alpha$ ($\frac{2}{5}$, $\frac{1}{2}$) can obtain a higher speedup but may lead to a slight loss in accuracy (See section \ref{sec:freeze_alpha_setting}). Note that the model size of BERT (24 layers) is larger than ViT-B/16 (12 layers), thus it takes more time for communication (see Section \ref{sec:communication_cost} for details).

\begin{table}[h!]
\caption{Speedup for ViT and BERT Training}
\label{table:speedup_cv}
\begin{center}
\resizebox{\linewidth}{!}{
\begin{tabular}{lccccc}
\toprule
 & \multicolumn{2}{c}{\textbf{Baseline}} & \multicolumn{2}{c}{\textbf{PipeTransformer}} &   \\
\cmidrule(lr){2-3}
\cmidrule(lr){4-5}
\multirow{2}{*}{Dataset}& \multirow{2}{*}{Accuracy} & Training & \multirow{2}{*}{Accuracy} & Training & Training \\
& & time & & time  & Speedup \\
\midrule
ImageNet    & 80.83 $\pm$ 0.05 & 26h 30m & 82.18 $\pm$ 0.32 & 9h 21m & \textbf{\large{2.83}} $\times$\\
CIFAR-100    & 91.21 $\pm$ 0.07 & 35m 6s & 91.33 $\pm$ 0.05 & 12m 23s & 2.44 $\times$\\
SQuAD 1.1    & 90.71 $\pm$ 0.18 & 5h 7m & 90.69 $\pm$ 0.23 & 2h 26m & 2.10 $\times$\\
\bottomrule
\end{tabular}
}
\end{center}
\begin{tablenotes}
      \footnotesize
      \item *Note: 1. the accuracy is the mean and variance of three independent runs with the same random seed; 2. the training time among different runs are relatively stable (the gap is less than 1 minute); 3. GLUE (SST-2)'s training time is relatively small, thus we mainly used it for debugging without reporting a few minutes result. 4. accuracy metric: ImageNet/CIFAR-100: top-1 accuracy; SQuAD: F1 score.  
\end{tablenotes}
\vspace{-0.5cm}
\end{table}

\subsection{Performance Analysis}

This section presents evaluation results and analyzes the performance of different components in \autopipe. More experimental results can be found in the Appendix.

\subsubsection{Speedup Breakdown}
\label{sec:speedup_breakdown}

\begin{figure}[h!]
\subfigure{{\includegraphics[width=0.95\linewidth]{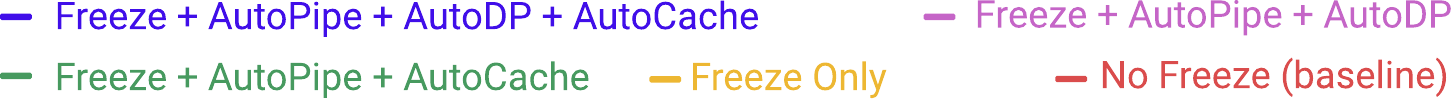}}}
\setcounter{subfigure}{0}
\setlength{\abovecaptionskip}{0pt}
\subfigure[\label{fig:throughput} Sample Throughput]
{{\includegraphics[width=0.48\linewidth]{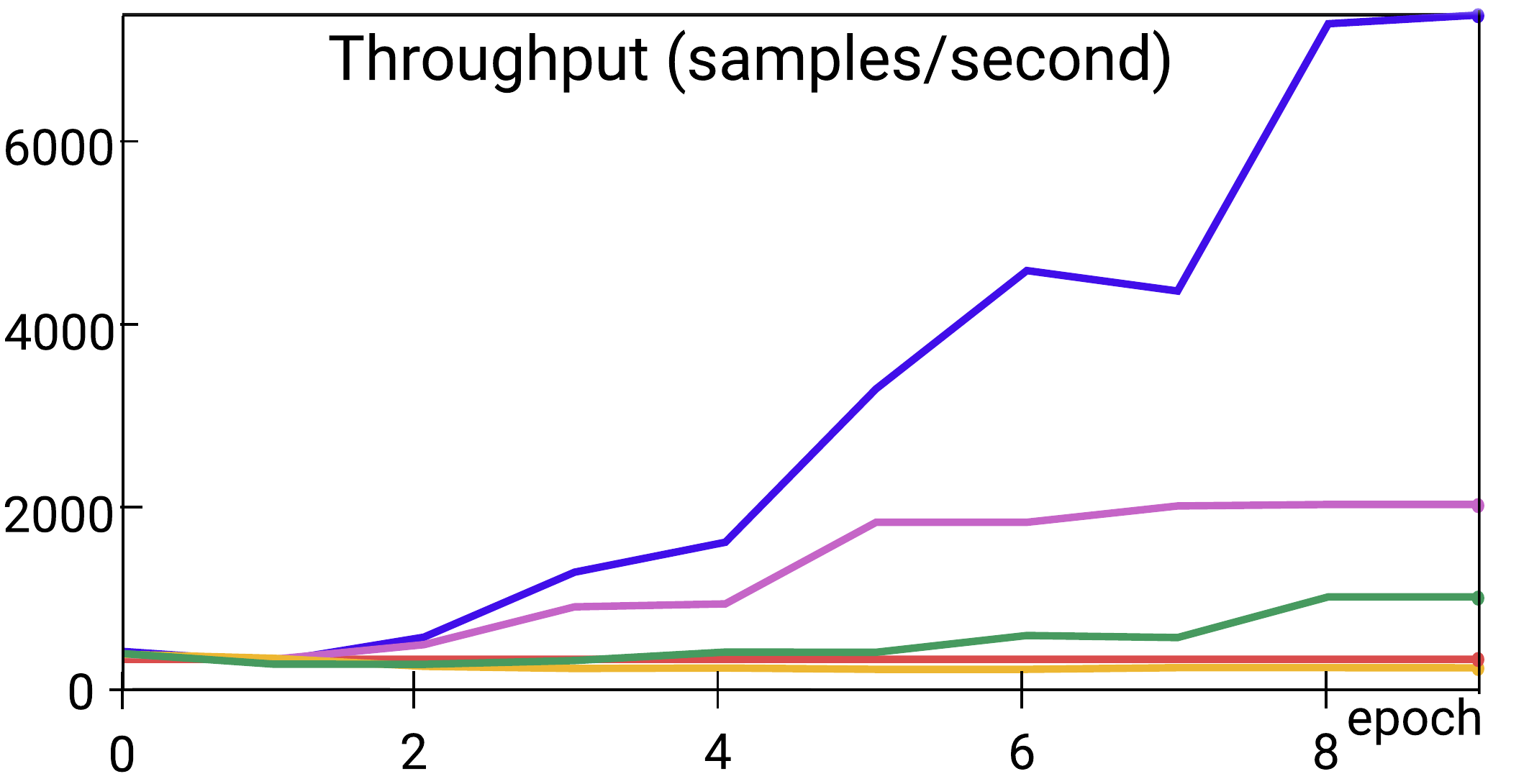}}}
\subfigure[Speedup Ratio Comparison]
{{\includegraphics[width=0.47\linewidth]{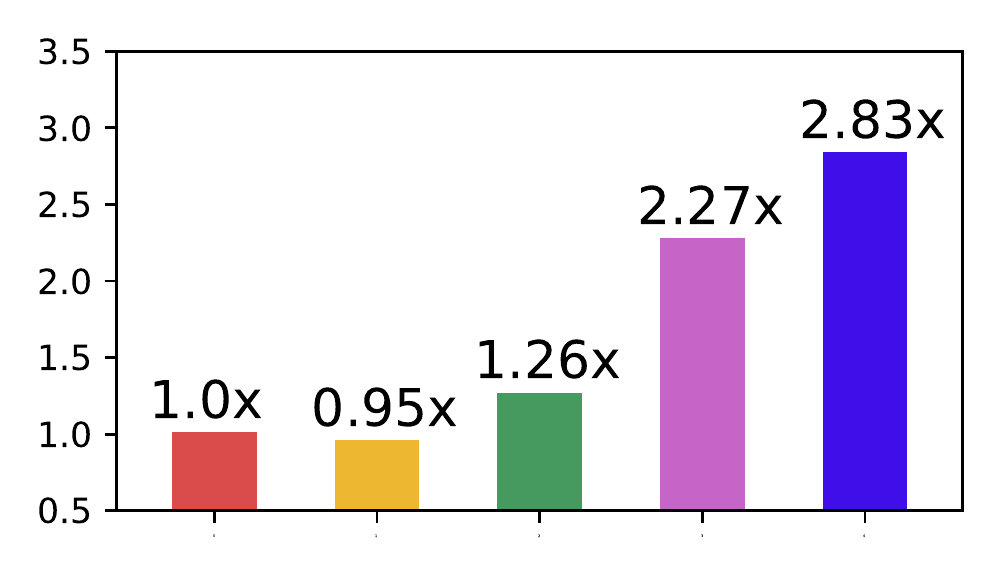}}}
\setlength{\belowcaptionskip}{-0.5cm}
\vspace{-0.3cm}
\caption{Speedup Breakdown (ViT on ImageNet)}
\label{fig:breakdown}
\vspace{-0.3cm}
\end{figure}


To understand the efficacy of all four components and their impacts on training speed, we experimented with different combinations and used their training sample throughput (samples/second) and speedup ratio as metrics. Results are illustrated in Figure \ref{fig:breakdown}. Key takeaways from these experimental results are: 
1. the main speedup is the result of elastic pipelining which is achieved through the joint use of \code{AutoPipe} and \code{AutoDP};
2. \code{AutoCache}'s contribution is amplified by \code{AutoDP};
3. freeze training alone without system-wise adjustment even downgrades the training speed (discussed in Section \ref{sec:auto_pipe}). We provide additional explanations of these results in the Appendix.

\begin{figure*}[htb]
\setcounter{subfigure}{0}
\subfigure[\label{fig:fig1a} Tuning $\alpha$ in Freeze Algorithm]
{{\includegraphics[width=0.35\textwidth]{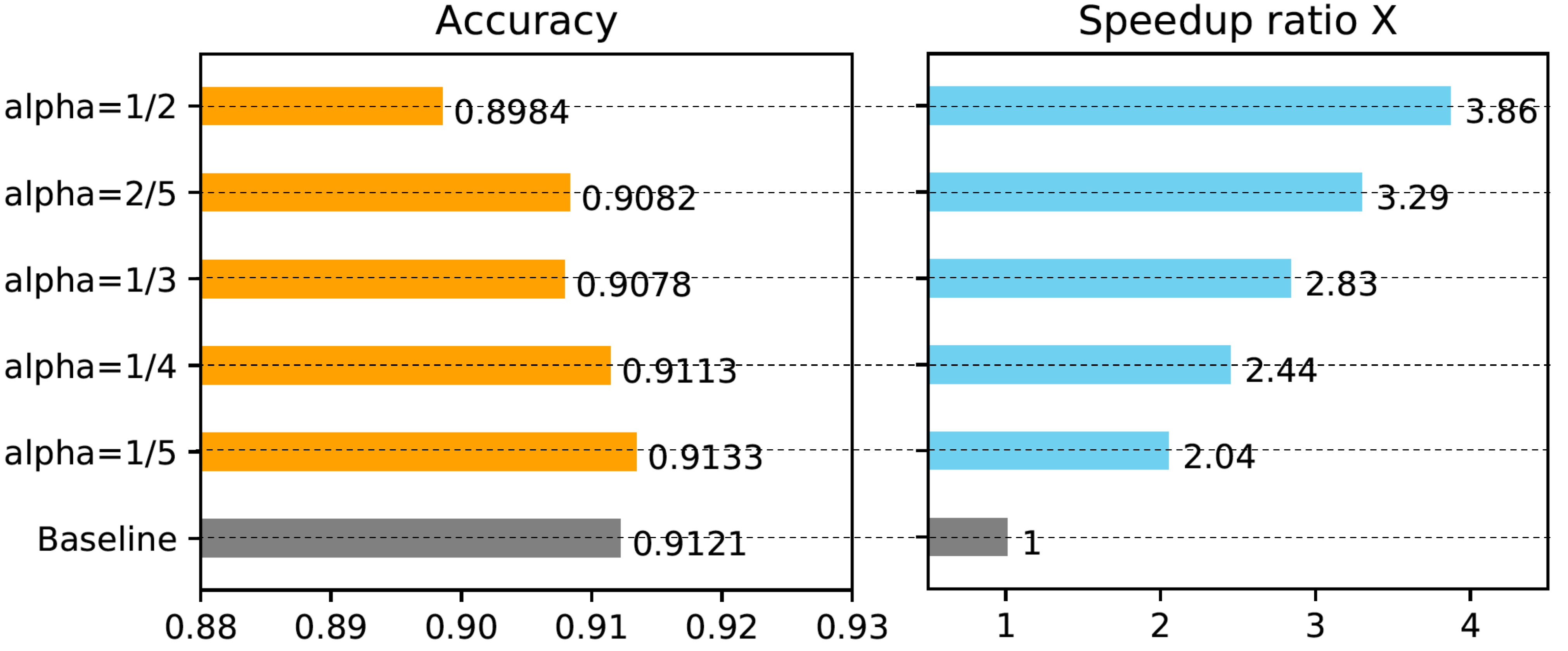}}}
\subfigure[Profiling Optimal Chunk Number]
{{\includegraphics[width=0.36\textwidth]{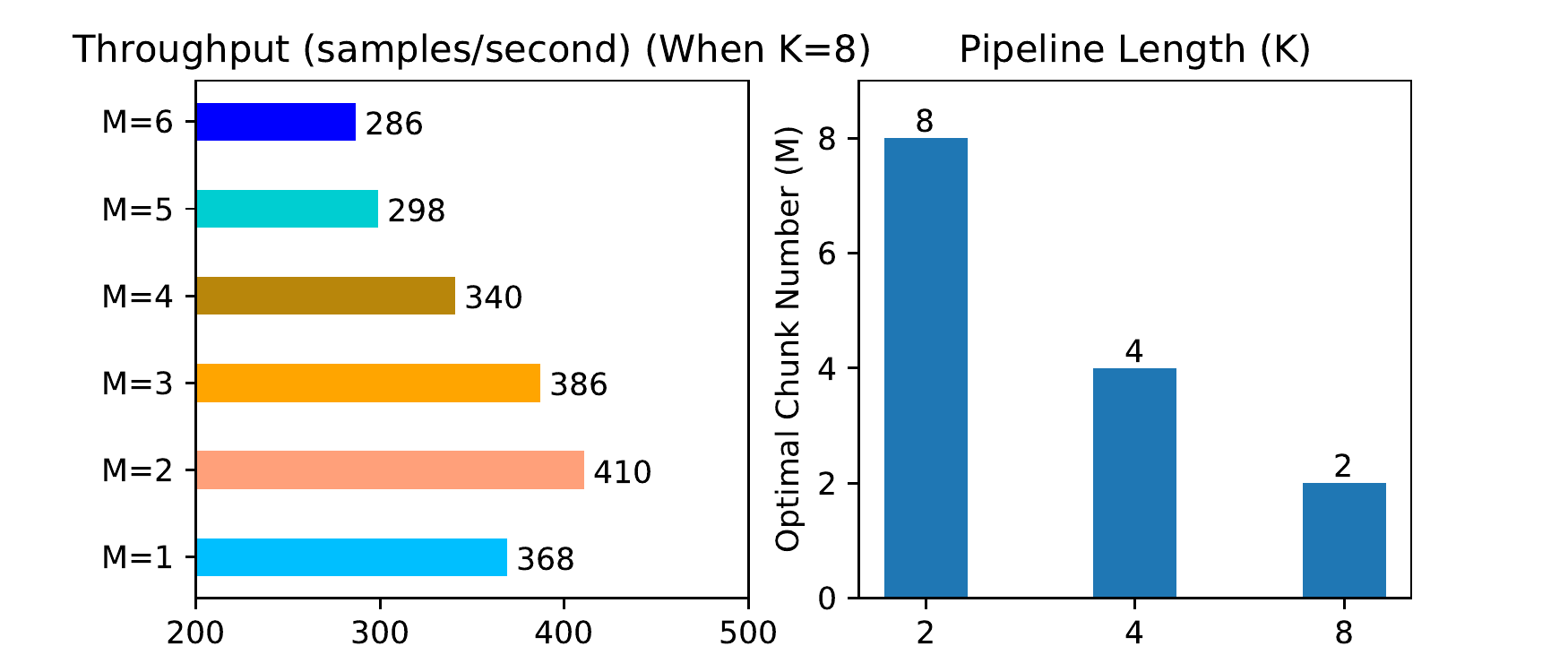}}}
\subfigure[\label{fig:fig1c} Timing of Caching]
{{\includegraphics[width=0.28\textwidth]{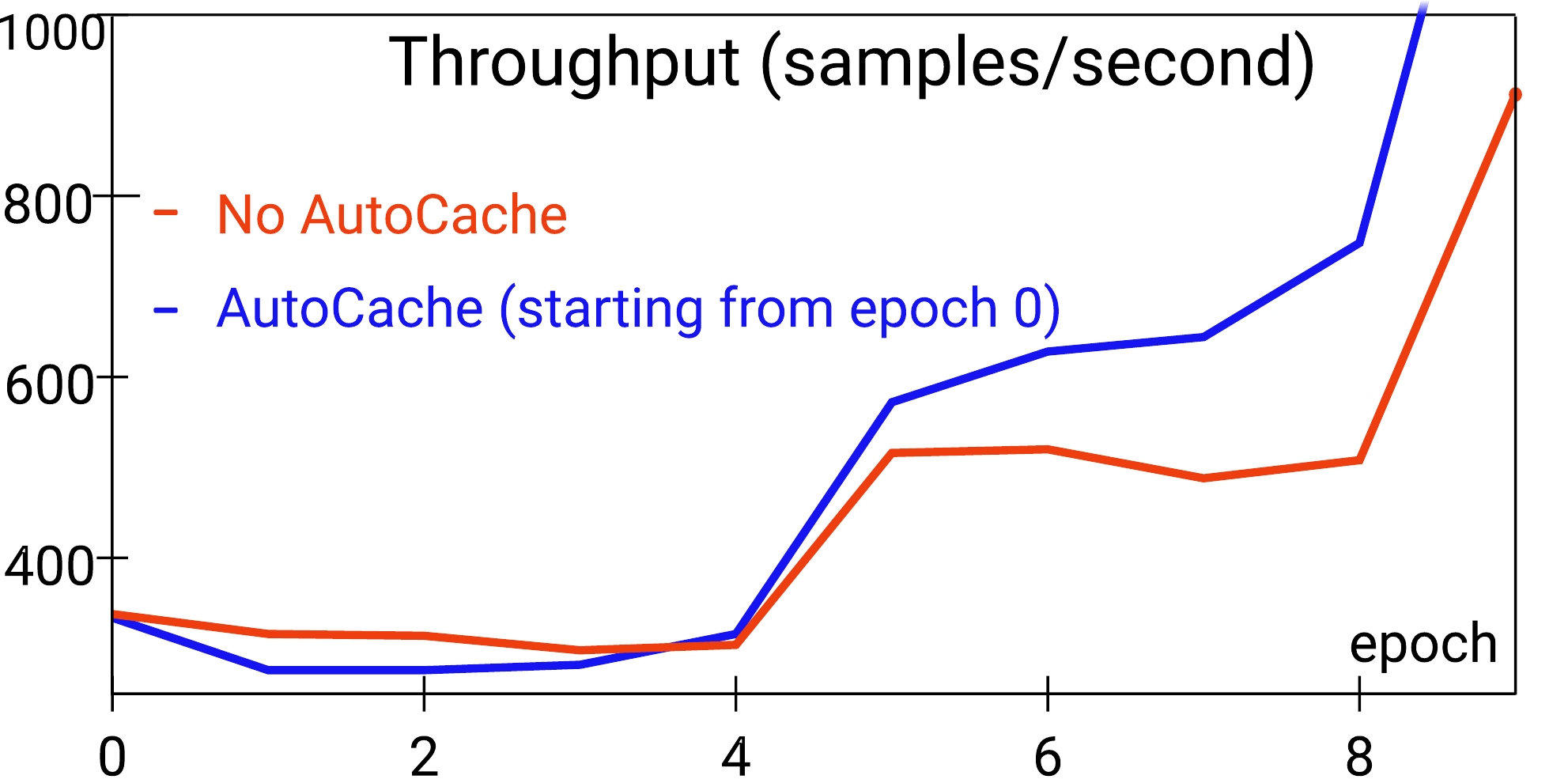}}}
\setlength{\belowcaptionskip}{-1cm}
\vspace{-0.5cm}
\caption{\textcolor{black}{Some Results of Performance Analysis}}
\label{fig:performance_analysis}
\vspace{-0.4cm}
 \end{figure*}
\subsubsection{Communication Cost}
\label{sec:communication_cost}



We also analyzed how communication and computation contribute to the overall training time. Since \code{PyTorch DDP} overlaps communication with computation, the time difference between a local training iteration and distributed training iteration does not faithfully represent the communication delay. Moreover, as DDP also organizes parameters into buckets and launches an \code{AllReduce} for each bucket, recording the start and finish time of overall communications also falls short, as there can be time gaps between buckets. To correctly measure DDP communication delay, we combined the DDP communication hook with \code{CUDAFuture} callback. More details of this measurement are documented in the Appendix. Key takeaways: 1. larger model cost more time on communication (BERT on SQuAD); 2. a higher cross-machine bandwidth can further speedup the training, especially for larger model.

\begin{table}[h!]
\vspace{-0.5cm}
\caption{Communication Cost v.s. Computational Cost}
\vspace{-0.2cm}
\label{table:communication_ratio}
\begin{center}
\resizebox{\linewidth}{!}{
\begin{threeparttable}
\begin{tabular}{lcccc}
\toprule
\multirow{2}{*}{\textbf{Dataset}} & \textbf{Overall} & \textbf{Communication} & \textbf{Computation} & \textbf{Communication}\\
 & \textbf{Cost} & \textbf{Cost} & \textbf{Cost} & \textbf{Cost Ratio}\\
\midrule
ImageNet & 9h 21m & 34m& 8h 47m & 5.9 \% \\
SQuAD & 2h 26m & 16m 33s & 2h 9m & 8.8\% \\
\bottomrule
\end{tabular}
\end{threeparttable}}
\end{center}
\vspace{-0.5cm}
\end{table}

\subsubsection{Tuning $\alpha$ in Freezing Algorithm}
\label{sec:freeze_alpha_setting}


We ran experiments to show how the $\alpha$ in the freeze algorithms influences training speed. The result clearly demonstrates that a larger $\alpha$ (excessive freeze) leads to a greater speedup but suffers from a slight performance degradation. In the case shown in Figure~\ref{fig:performance_analysis}(a), where $\alpha=1/5$, freeze training outperforms normal training and obtains a $2.04$-fold speedup. We provide more results in the Appendix.

\subsubsection{Optimal Chunks in elastic pipeline}


We profiled the optimal number of micro-batches $M$ for different pipeline lengths $K$. Results are summarized in Figure~\ref{fig:performance_analysis}(b). As we can see, different $K$ values lead to different optimal $M$, and the throughput gaps across different M values are large (as shown when $K=8$), which confirms the necessity of an anterior profiler in elastic pipelining.

\subsubsection{Understanding the Timing of Caching}
\label{sec:timing_of_caching}


To evaluate \code{AutoCache}, we compared the sample throughput of training that activates \code{AutoCache} from epoch $0$ (blue) with the training job without \code{AutoCache} (red). Figure \ref{fig:performance_analysis}(c) shows that enabling caching too early can slow down training, as caching can be more expensive than forward propagation on a small number of frozen layers. After freezing more layers, caching activations clearly outperforms the corresponding forward propagation. As a result, \code{AutoCache} uses a profiler to determine the proper timing to enable caching. In our system, for ViT (12 layers), caching starts from 3 frozen layers, while for BERT (24 layers), caching starts from 5 frozen layers.

\section{Related Works}

\code{PipeTransformer} combines pipeline parallelism~\cite{gpipe, pipedream, hetpipe} and data parallelism~\cite{ddp}. Both techniques have been extensively studied in prior work. GPipe~\cite{gpipe} parallelizes micro-batches within a mini-batch and enforces synchronizations between consecutive mini-batches.
The synchronization barrier creates execution bubbles and it exacerbates if the model spans across more devices. PipeDream~\cite{pipedream} and HetPipe~\cite{hetpipe} remove or mitigate execution bubbles by allowing a configurable amount of staleness. Although evaluations show that models can still converge with high accuracy, it breaks the mathematical equivalence to local training. \code{PipeTransformer} builds on top of PyTorch pipeline parallel and distributed data-parallel APIs~\cite{ddp}.
Compared to prior solutions, \code{PipeTransformer} reduces the size of bubbles during training by dynamically packing the active layers into fewer GPUs. Moreover, the communication overhead for data-parallel training, which is the dominant source of delay, also drops when the active model size shrinks. 

\section{Discussion}
We defer the discussion section to the Appendix, where we discuss \textit{pretraining v.s. fine-tuning}, \textit{designing better freeze algorithms}, and the  \textit{versatility} of our approach.






\section{Conclusion}

This paper proposes \autopipe, a holistic solution that combines elastic pipeline-parallel and data-parallel for distributed training. More specifically, \autopipe\ incrementally freezes layers in the pipeline, packs remaining active layers into fewer GPUs, and forks more pipeline replicas to increase the data-parallel width. Evaluations on ViT and BERT models show that compared to the state-of-the-art baseline, \autopipe\ attains up to $2.83\times$ speedups without accuracy loss.



\nocite{langley00}

\bibliography{PipeTransformer}

\begin{thebibliography}{24}
\providecommand{\natexlab}[1]{#1}
\providecommand{\url}[1]{\texttt{#1}}
\expandafter\ifx\csname urlstyle\endcsname\relax
  \providecommand{\doi}[1]{doi: #1}\else
  \providecommand{\doi}{doi: \begingroup \urlstyle{rm}\Url}\fi

\bibitem[Brown et~al.(2020)Brown, Mann, Ryder, Subbiah, Kaplan, Dhariwal,
  Neelakantan, Shyam, Sastry, Askell, et~al.]{gpt3}
Brown, T.~B., Mann, B., Ryder, N., Subbiah, M., Kaplan, J., Dhariwal, P.,
  Neelakantan, A., Shyam, P., Sastry, G., Askell, A., et~al.
\newblock Language models are few-shot learners.
\newblock \emph{arXiv preprint arXiv:2005.14165}, 2020.

\bibitem[Devlin et~al.(2018)Devlin, Chang, Lee, and Toutanova]{devlin2018bert}
Devlin, J., Chang, M.-W., Lee, K., and Toutanova, K.
\newblock Bert: Pre-training of deep bidirectional transformers for language
  understanding.
\newblock \emph{arXiv preprint arXiv:1810.04805}, 2018.

\bibitem[Dosovitskiy et~al.(2020)Dosovitskiy, Beyer, Kolesnikov, Weissenborn,
  Zhai, Unterthiner, Dehghani, Minderer, Heigold, Gelly,
  et~al.]{dosovitskiy2020image}
Dosovitskiy, A., Beyer, L., Kolesnikov, A., Weissenborn, D., Zhai, X.,
  Unterthiner, T., Dehghani, M., Minderer, M., Heigold, G., Gelly, S., et~al.
\newblock An image is worth 16x16 words: Transformers for image recognition at
  scale.
\newblock \emph{arXiv preprint arXiv:2010.11929}, 2020.

\bibitem[He et~al.(2016)He, Zhang, Ren, and Sun]{he2016deep}
He, K., Zhang, X., Ren, S., and Sun, J.
\newblock Deep residual learning for image recognition.
\newblock In \emph{Proceedings of the IEEE conference on computer vision and
  pattern recognition}, pp.\  770--778, 2016.

\bibitem[Huang et~al.(2018)Huang, Cheng, Bapna, Firat, Chen, Chen, Lee, Ngiam,
  Le, Wu, et~al.]{huang2018gpipe}
Huang, Y., Cheng, Y., Bapna, A., Firat, O., Chen, M.~X., Chen, D., Lee, H.,
  Ngiam, J., Le, Q.~V., Wu, Y., et~al.
\newblock Gpipe: Efficient training of giant neural networks using pipeline
  parallelism.
\newblock \emph{arXiv preprint arXiv:1811.06965}, 2018.

\bibitem[Huang et~al.(2019)Huang, Cheng, Bapna, Firat, Chen, Chen, Lee, Ngiam,
  Le, Wu, and Chen]{gpipe}
Huang, Y., Cheng, Y., Bapna, A., Firat, O., Chen, D., Chen, M., Lee, H., Ngiam,
  J., Le, Q.~V., Wu, Y., and Chen, z.
\newblock Gpipe: Efficient training of giant neural networks using pipeline
  parallelism.
\newblock In Wallach, H., Larochelle, H., Beygelzimer, A., d\textquotesingle
  Alch\'{e}-Buc, F., Fox, E., and Garnett, R. (eds.), \emph{Advances in Neural
  Information Processing Systems}, volume~32, pp.\  103--112. Curran
  Associates, Inc., 2019.

\bibitem[Jiang et~al.(2020)Jiang, Zhu, Lan, Yi, Cui, and Guo]{byteps}
Jiang, Y., Zhu, Y., Lan, C., Yi, B., Cui, Y., and Guo, C.
\newblock A unified architecture for accelerating distributed {DNN} training in
  heterogeneous gpu/cpu clusters.
\newblock In \emph{14th {USENIX} Symposium on Operating Systems Design and
  Implementation ({OSDI} 20)}, pp.\  463--479. {USENIX} Association, November
  2020.
\newblock ISBN 978-1-939133-19-9.
\newblock URL
  \url{https://www.usenix.org/conference/osdi20/presentation/jiang}.

\bibitem[Kim et~al.(2020)Kim, Lee, Jeong, Baek, Yoon, Kim, Lim, and
  Kim]{kim2020torchgpipe}
Kim, C., Lee, H., Jeong, M., Baek, W., Yoon, B., Kim, I., Lim, S., and Kim, S.
\newblock torchgpipe: On-the-fly pipeline parallelism for training giant
  models.
\newblock \emph{arXiv preprint arXiv:2004.09910}, 2020.

\bibitem[Kim et~al.(2019)Kim, Yu, Park, Cho, Jeong, Ha, Lee, Jeong, and
  Chun]{parallax}
Kim, S., Yu, G.-I., Park, H., Cho, S., Jeong, E., Ha, H., Lee, S., Jeong,
  J.~S., and Chun, B.-G.
\newblock Parallax: Sparsity-aware data parallel training of deep neural
  networks.
\newblock In \emph{Proceedings of the Fourteenth EuroSys Conference 2019}, pp.\
   1--15, 2019.

\bibitem[Lepikhin et~al.(2020)Lepikhin, Lee, Xu, Chen, Firat, Huang, Krikun,
  Shazeer, and Chen]{gshard}
Lepikhin, D., Lee, H., Xu, Y., Chen, D., Firat, O., Huang, Y., Krikun, M.,
  Shazeer, N., and Chen, Z.
\newblock Gshard: Scaling giant models with conditional computation and
  automatic sharding.
\newblock \emph{arXiv preprint arXiv:2006.16668}, 2020.

\bibitem[Li et~al.(2014)Li, Andersen, Park, Smola, Ahmed, Josifovski, Long,
  Shekita, and Su]{ps}
Li, M., Andersen, D.~G., Park, J.~W., Smola, A.~J., Ahmed, A., Josifovski, V.,
  Long, J., Shekita, E.~J., and Su, B.-Y.
\newblock Scaling distributed machine learning with the parameter server.
\newblock In \emph{11th $\{$USENIX$\}$ Symposium on Operating Systems Design
  and Implementation ($\{$OSDI$\}$ 14)}, pp.\  583--598, 2014.

\bibitem[Li et~al.(2020)Li, Zhao, Varma, Salpekar, Noordhuis, Li, Paszke,
  Smith, Vaughan, Damania, et~al.]{ddp}
Li, S., Zhao, Y., Varma, R., Salpekar, O., Noordhuis, P., Li, T., Paszke, A.,
  Smith, J., Vaughan, B., Damania, P., et~al.
\newblock Pytorch distributed: Experiences on accelerating data parallel
  training.
\newblock \emph{Proceedings of the VLDB Endowment}, 13\penalty0 (12), 2020.

\bibitem[Morcos et~al.(2018)Morcos, Raghu, and Bengio]{NIPS2018_7815}
Morcos, A., Raghu, M., and Bengio, S.
\newblock Insights on representational similarity in neural networks with
  canonical correlation.
\newblock In Bengio, S., Wallach, H., Larochelle, H., Grauman, K.,
  Cesa-Bianchi, N., and Garnett, R. (eds.), \emph{Advances in Neural
  Information Processing Systems 31}, pp.\  5732--5741. Curran Associates,
  Inc., 2018.
\newblock URL
  \url{http://papers.nips.cc/paper/7815-insights-on-representational-similarity-in-neural-networks-with-canonical-correlation.pdf}.

\bibitem[Narayanan et~al.(2019)Narayanan, Harlap, Phanishayee, Seshadri,
  Devanur, Ganger, Gibbons, and Zaharia]{pipedream}
Narayanan, D., Harlap, A., Phanishayee, A., Seshadri, V., Devanur, N.~R.,
  Ganger, G.~R., Gibbons, P.~B., and Zaharia, M.
\newblock Pipedream: Generalized pipeline parallelism for dnn training.
\newblock In \emph{Proceedings of the 27th ACM Symposium on Operating Systems
  Principles}, SOSP '19, pp.\  1–15, New York, NY, USA, 2019. Association for
  Computing Machinery.
\newblock ISBN 9781450368735.
\newblock \doi{10.1145/3341301.3359646}.

\bibitem[Park et~al.(2020)Park, Yun, Yi, Nguyen, Lee, Choi, Noh, and
  ri~Choi]{hetpipe}
Park, J.~H., Yun, G., Yi, C.~M., Nguyen, N.~T., Lee, S., Choi, J., Noh, S.~H.,
  and ri~Choi, Y.
\newblock Hetpipe: Enabling large {DNN} training on (whimpy) heterogeneous
  {GPU} clusters through integration of pipelined model parallelism and data
  parallelism.
\newblock In \emph{2020 {USENIX} Annual Technical Conference ({USENIX} {ATC}
  20)}, pp.\  307--321. {USENIX} Association, July 2020.
\newblock ISBN 978-1-939133-14-4.
\newblock URL \url{https://www.usenix.org/conference/atc20/presentation/park}.

\bibitem[Paszke et~al.(2019)Paszke, Gross, Massa, Lerer, Bradbury, Chanan,
  Killeen, Lin, Gimelshein, Antiga, et~al.]{paszke2019pytorch}
Paszke, A., Gross, S., Massa, F., Lerer, A., Bradbury, J., Chanan, G., Killeen,
  T., Lin, Z., Gimelshein, N., Antiga, L., et~al.
\newblock Pytorch: An imperative style, high-performance deep learning library.
\newblock \emph{arXiv preprint arXiv:1912.01703}, 2019.

\bibitem[Raghu et~al.(2017)Raghu, Gilmer, Yosinski, and
  Sohl-Dickstein]{Raghu2017SVCCASV}
Raghu, M., Gilmer, J., Yosinski, J., and Sohl-Dickstein, J.
\newblock Svcca: Singular vector canonical correlation analysis for deep
  learning dynamics and interpretability.
\newblock In \emph{NIPS}, 2017.

\bibitem[Rajbhandari et~al.(2019)Rajbhandari, Rasley, Ruwase, and He]{zero}
Rajbhandari, S., Rasley, J., Ruwase, O., and He, Y.
\newblock Zero: Memory optimization towards training a trillion parameter
  models.
\newblock \emph{arXiv preprint arXiv:1910.02054}, 2019.

\bibitem[Shazeer et~al.(2018)Shazeer, Cheng, Parmar, Tran, Vaswani,
  Koanantakool, Hawkins, Lee, Hong, Young, Sepassi, and Hechtman]{meshtf}
Shazeer, N., Cheng, Y., Parmar, N., Tran, D., Vaswani, A., Koanantakool, P.,
  Hawkins, P., Lee, H., Hong, M., Young, C., Sepassi, R., and Hechtman, B.
\newblock Mesh-tensorflow: Deep learning for supercomputers.
\newblock In Bengio, S., Wallach, H., Larochelle, H., Grauman, K.,
  Cesa-Bianchi, N., and Garnett, R. (eds.), \emph{Advances in Neural
  Information Processing Systems}, volume~31, pp.\  10414--10423. Curran
  Associates, Inc., 2018.

\bibitem[Shen et~al.(2020)Shen, Baevski, Morcos, Keutzer, Auli, and
  Kiela]{reservoir}
Shen, S., Baevski, A., Morcos, A.~S., Keutzer, K., Auli, M., and Kiela, D.
\newblock Reservoir transformer.
\newblock \emph{arXiv preprint arXiv:2012.15045}, 2020.

\bibitem[Shoeybi et~al.(2019)Shoeybi, Patwary, Puri, LeGresley, Casper, and
  Catanzaro]{megatron}
Shoeybi, M., Patwary, M., Puri, R., LeGresley, P., Casper, J., and Catanzaro,
  B.
\newblock Megatron-lm: Training multi-billion parameter language models using
  model parallelism.
\newblock \emph{arXiv preprint arXiv:1909.08053}, 2019.

\bibitem[Tan \& Le(2019)Tan and Le]{tan2019efficientnet}
Tan, M. and Le, Q.
\newblock Efficientnet: Rethinking model scaling for convolutional neural
  networks.
\newblock In \emph{International Conference on Machine Learning}, pp.\
  6105--6114. PMLR, 2019.

\bibitem[Vaswani et~al.(2017)Vaswani, Shazeer, Parmar, Uszkoreit, Jones, Gomez,
  Kaiser, and Polosukhin]{vaswani2017attention}
Vaswani, A., Shazeer, N., Parmar, N., Uszkoreit, J., Jones, L., Gomez, A.~N.,
  Kaiser, L., and Polosukhin, I.
\newblock Attention is all you need.
\newblock \emph{arXiv preprint arXiv:1706.03762}, 2017.

\bibitem[Xiong et~al.(2020)Xiong, Yang, He, Zheng, Zheng, Xing, Zhang, Lan,
  Wang, and Liu]{xiong2020layer}
Xiong, R., Yang, Y., He, D., Zheng, K., Zheng, S., Xing, C., Zhang, H., Lan,
  Y., Wang, L., and Liu, T.
\newblock On layer normalization in the transformer architecture.
\newblock In \emph{International Conference on Machine Learning}, pp.\
  10524--10533. PMLR, 2020.

\end{thebibliography}
\bibliographystyle{icml2021}

\newpage
\clearpage
\appendix

\section*{Appendix Outline}

This Appendix provides background and preliminaries, more details of four components, additional experimental details and results, and discussions. The organization is as follows:

\paragraph{Background and Preliminaries.} Appendix \ref{sec_app:bg_pre} provides the introduction for Transformer models, freeze training, pipeline parallelism, data parallelism, and hybrid of pipeline parallelism and data parallelism. This section serves as the required knowledge to understand \autopipe.

\paragraph{More Details of \code{Freeze Algorithm}, \code{AutoPipe}, \code{AutoDP}, \code{AutoCache}.} Appendix \ref{app:freeze} explains more details of design motivation for freeze training algorithm and shows details of the deviation; Appendix \ref{appendix:load_balance} provides more analysis to understand the design choice of \code{AutoPipe}; Appendix \ref{app_sec:autodp} contains more details of \code{AutoDP}, including the dataset redistributing, and comparing another way to skip frozen parameters; Appendix \ref{appendix:caching} introduces additional details for \code{AutoCache}.

\paragraph{More Experimental Results and Details.} In Appendix \ref{app_sec:experiments}, we provide hyper-parameters and more experimental results. Especially, we provide more details of speedup breakdown in \ref{app_sec:speedup_breakdown}.

\paragraph{Discussion.} In Appendix \ref{appsec:discussion}, we will discuss pretraining v.s. fine-tuning, designing better freeze algorithms, and the versatility of our approach.

\section{Background and Preliminaries}
\label{sec_app:bg_pre}

\subsection{Transformer Models: ViT and BERT}

\begin{figure}[h!]
    \centering
    \includegraphics[width = 1 \linewidth]{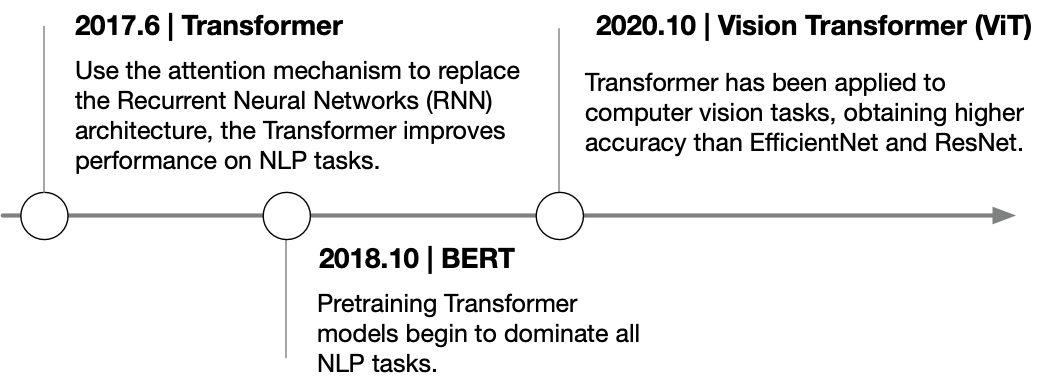}
    \caption{Evolution of Transformer Models.}
    \label{fig:transformer_history}
\end{figure}

\begin{figure}[h!]
    \centering
    \includegraphics[width = 0.6 \linewidth]{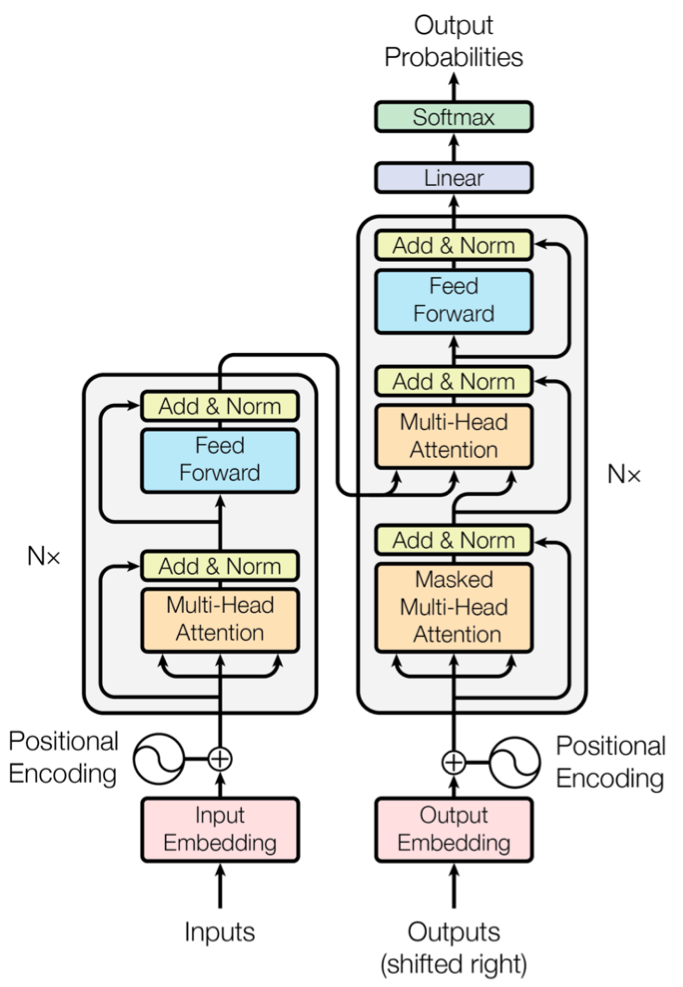}
    \caption{Transformer Model Architecture \cite{vaswani2017attention}}
    \label{fig:transformer_architecture}
\end{figure}

\paragraph{Transformer.} The Transformer model originates from the Natural Language Processing (NLP) community. It replaces the recurrent neural network (RNN) using a \textit{self-attention} mechanism which relates different positions of a single sequence in order to compute a representation of the sequence. The transformer model has an encoder-decoder structure which is a classical structure for sequence modeling. The encoder maps an input sequence of symbol representations $\left(x_{1}, \ldots, x_{n}\right)$ to a sequence of continuous representations $\mathbf{z}=\left(z_{1}, \ldots, z_{n}\right)$. Given $\mathbf{z}$, the decoder then generates an output sequence $\left(y_{1}, \ldots, y_{m}\right)$ of symbols one element at a time. As shown in Figure \ref{fig:transformer_architecture}, the Transformer follows this overall architecture using stacked self-attention and point-wise, fully connected layers for both the encoder (left) and decoder (right). To better understand this architecture, we refer readers to the tutorial ``The Annotated Transformer'' \footnote{\url{ http://nlp.seas.harvard.edu/2018/04/03/attention.html}}.

\paragraph{BERT (ViT).} BERT \cite{devlin2018bert}, which stands for Bidirectional Encoder Representations from Transformers, simply stacks multiple Transformer encoders (also called the Transformer layer, Figure \ref{fig:transformer_architecture}, left). $\text { BERT }_{\text {BASE }}$ has 12 Transformer layers, and its total number of parameters is 110M. $\text { BERT }_{\text {LARGE }}$ has 24 Transformer layers, and its total number of parameters is 340M. BERT is pre-trained using unsupervised tasks (masked language model, and next sentence prediction) and then fine-tuned to various NLP tasks such as text classification and question answering.

\begin{figure}[h!]
    \centering
    \includegraphics[width = 0.8 \linewidth]{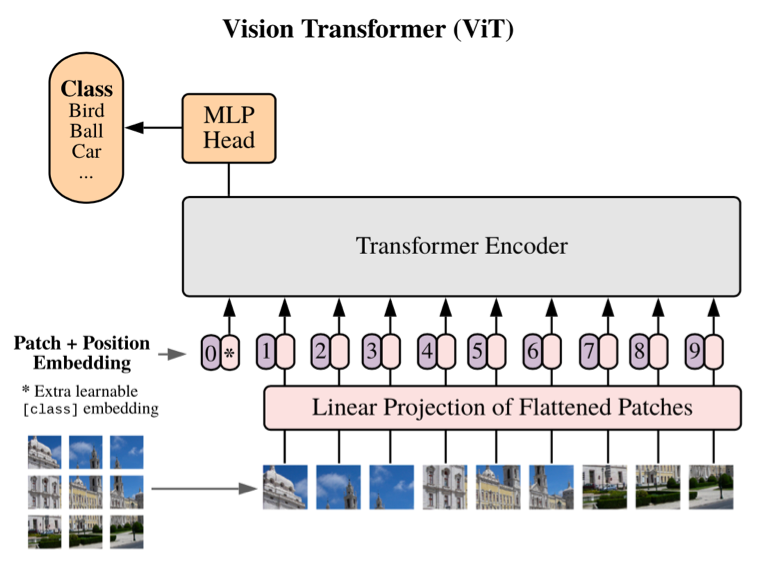}
    \caption{Vision Transformer \cite{dosovitskiy2020image}}
    \label{fig:vit_arch}
\end{figure}
\paragraph{Vision Transformer (ViT).} ViT \cite{dosovitskiy2020image} attains excellent results compared to state-of-the-art convolutional networks. Its architecture is shown in Figure \ref{fig:vit_arch}. It splits an image into fixed-size patches, linearly embeds each of them, adds position embeddings, and feeds the resulting sequence of vectors to a Transformer encoder. Similar to BERT, the Transformer encode repeats multiple layers. 

\begin{figure}[h!]
    \centering
    \includegraphics[width = 1.0 \linewidth]{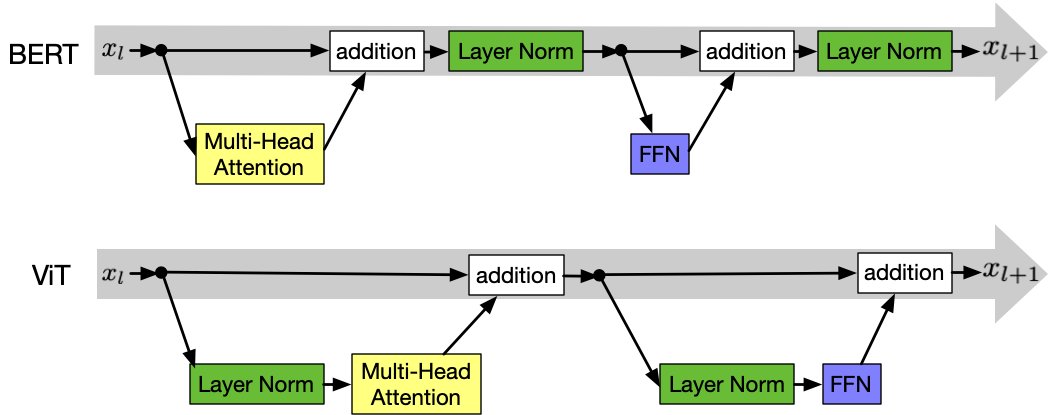}
    \caption{Comparison of Transform in BERT and ViT}
    \label{fig:compare_vit_and_BERT}
\end{figure}

\paragraph{Model Architecture Comparison.} Note that ViT and BERT's Transformer encoder places layer normalization in different locations. To understand the differences between these two architectures, please refer to the analysis in \cite{xiong2020layer}. Due to this slight difference, our \autopipe\ source code implements the model partition of these two architectures separately.

\subsection{Freeze Training.} The concept of freeze training is first proposed by \cite{Raghu2017SVCCASV}, which provides a posterior algorithm, named SVCCA (Singular Vector Canonical Correlation Analysis), to compare two representations. SVCCA can compare the representation at a layer at different points during training to its final representation and find that lower layers tend to converge faster than higher layers. This means that not all layers need to be trained through training. We can save computation and prevent overfitting by consecutively freezing layers. However, SVCCA has to take the entire dataset as its input, which does not fit an on-the-fly analysis. This drawback motivates us to design an adaptive on the fly freeze algorithm.

\subsection{Pipeline Parallelism}
\label{app:pipelining}

\begin{figure}[h!]
    \centering
    \includegraphics[width = 1.0 \linewidth]{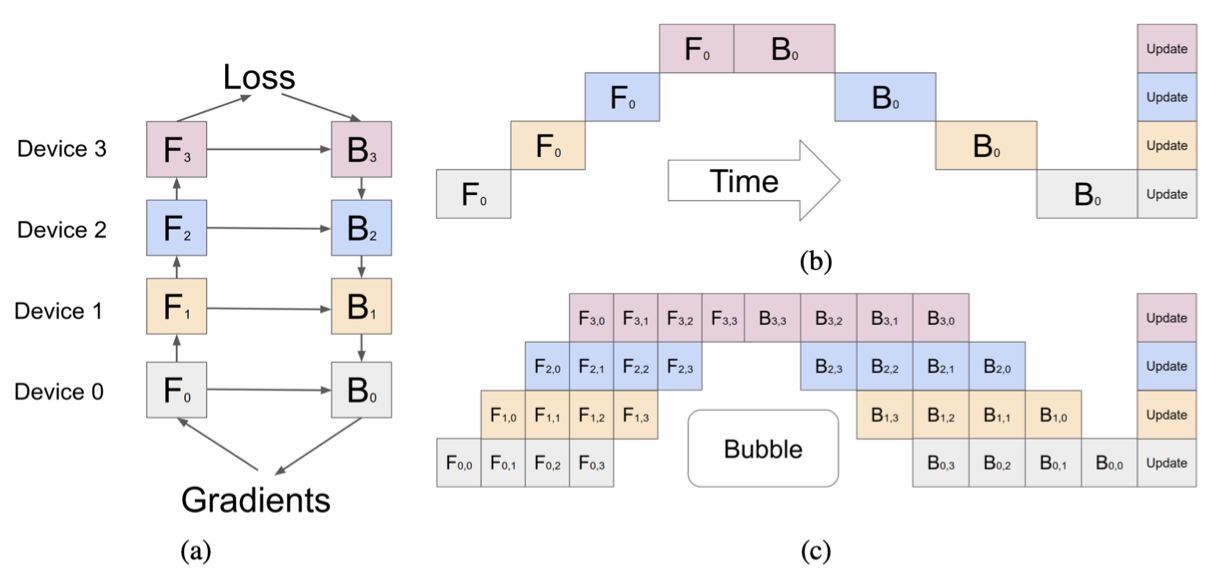}
    \caption{GPipe \cite{huang2018gpipe}}
    \label{fig:gpipe}
\end{figure}

In \autopipe, we reuse \texttt{GPipe} as the baseline. \texttt{GPipe} is a pipeline parallelism library that can divide different sub-sequences of layers to separate accelerators, which provides the flexibility of scaling a variety of different networks to gigantic sizes efficiently. The key design in \texttt{GPipe} is that it splits the mini-batch into $M$ micro-batches, which can train faster than naive model parallelism (as shown in Figure \ref{fig:gpipe}(b). However, as illustrated in Figure \ref{fig:gpipe}(c), micro-batches still cannot thoroughly avoid bubble overhead (some idle time per accelerator). \texttt{GPipe} empirically demonstrates that the bubble overhead is negligible when $M \geq 4 \times K$. Different from \texttt{GPipe}, \autopipe\ has an elastic pipelining parallelism in which $K$ and pipeline number are dynamic during the training.

\subsection{Data Parallelism}
\label{app:ddp}

\begin{figure}[h!]
    \centering
    \includegraphics[width=0.7 \linewidth]{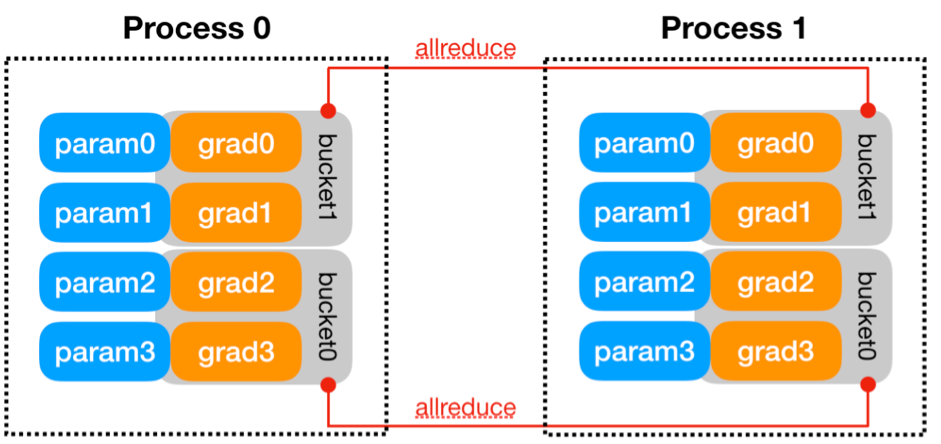}
    \caption{\code{PyTorch DDP} Bucket-based \code{AllReduce}}
    \label{fig:ddp_bucet}
\end{figure}
In \texttt{PyTorch DDP} \cite{ddp}, to improve communication efficiency, gradients are organized into buckets, and \code{AllReduce} is operated on one bucket at a time. The mapping from parameter gradients to buckets is determined at the construction time, based on the bucket size limit and parameter sizes. Model parameters are allocated into buckets in (roughly) the reverse order of \code{Model.parameters()} from the given model. Reverse order is used because \texttt{DDP} expects gradients to be ready during the backward pass in approximately that order. Figure \ref{fig:ddp_bucet} shows an example. Note that, grad0 and grad1 are in bucket1, and the other two gradients are in bucket0. With this bucket design, \texttt{DDP} can overlap part of the communication time with the computation time of backward propagation. 

\subsection{Hybrid of Pipeline Parallelism and Data Parallelism}
\label{sec:hybrid}

\begin{figure}[h!]
    \centering
    \includegraphics[width = 0.8 \linewidth]{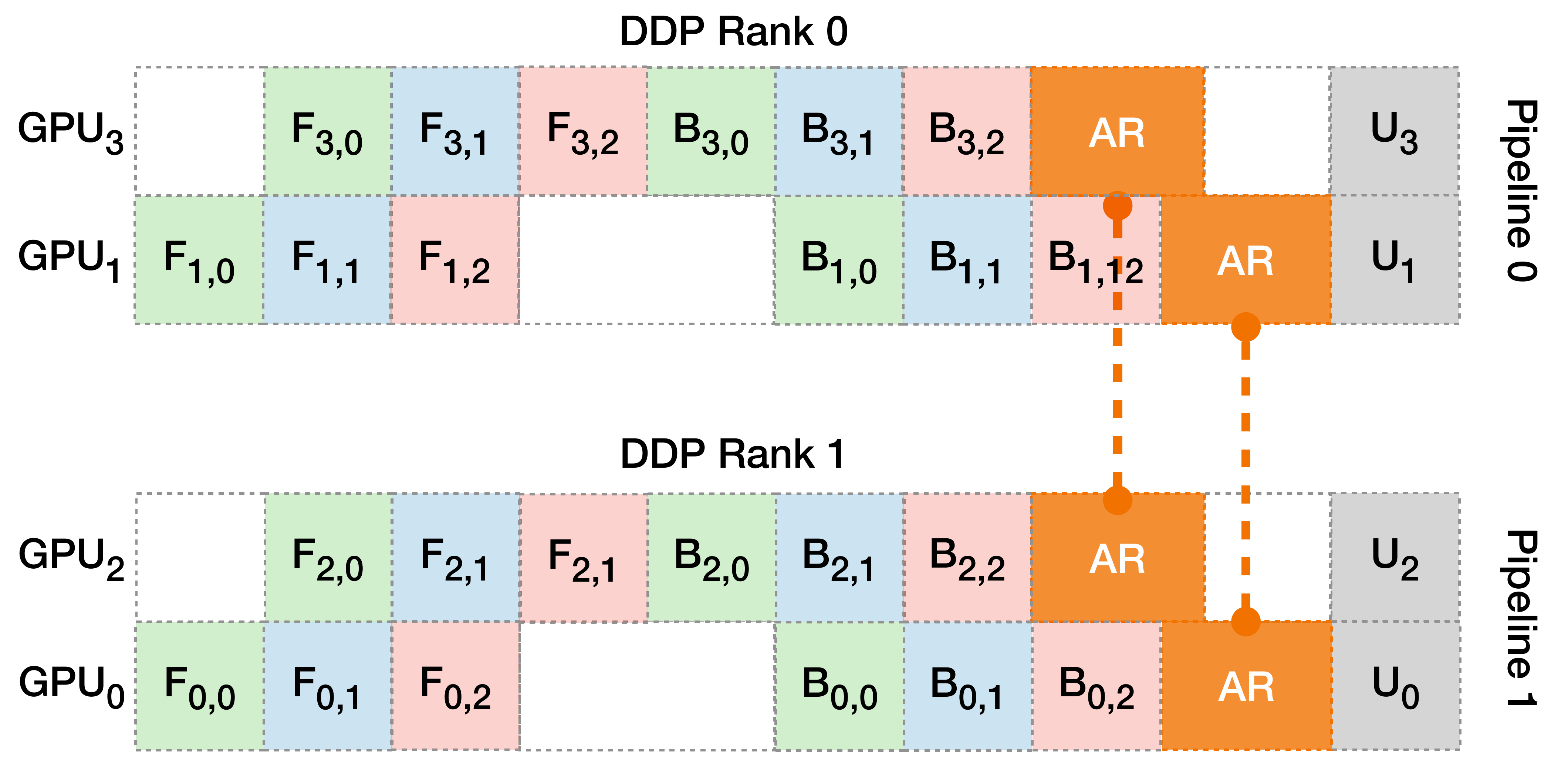}
    \caption{Illustration for Hybrid of Pipeline-parallel and Data-parallel}
    \label{fig:hybrid}
\end{figure}

To understand the hybrid of pipeline parallelism and data parallelism, we illustrate the training process in Figure \ref{fig:hybrid}. 
This example is hybrid two-way data parallelism and two-stage pipeline parallelism: pipeline 0 has two partitions, using GPU 1 and 3; pipeline 1 also has two partitions, using GPU 0 and 2; two pipelines are synchronized by data parallelism. 
Each batch of training data is divided into micro-batches that can be processed in parallel by the pipeline partitions. Once a partition completes the forward pass for a micro-batch, the activation memory is communicated to the pipeline's next partition. Similarly, as the next partition completes its backward pass on a micro-batch, the gradient with respect to the activation is communicated backward through the pipeline. Each backward pass accumulates gradients locally. Subsequently, all data parallel groups perform \code{AllReduce} on gradients.

 In this example, to simplify the figure, we assume that the bucket size is large enough to fit all gradients on a single device. That is to say, DDP uses one bucket per device, resulting in two \code{AllReduce} operations. Note that, since \code{AllReduce} can start as soon as gradients in corresponding buckets become ready. In this example, DDP launches \code{AllReduce} on GPU 1 and 3 immediately after $B_{3,1}$ and $B_{1, 1}$, without waiting for the rest of backward computation. Lastly, the optimizer updates the model weights. 

\section{More Details of Freeze Algorithm}
\label{app:freeze}

\begin{figure}[h!]
    \centering
    \includegraphics[width = 0.8 \linewidth]{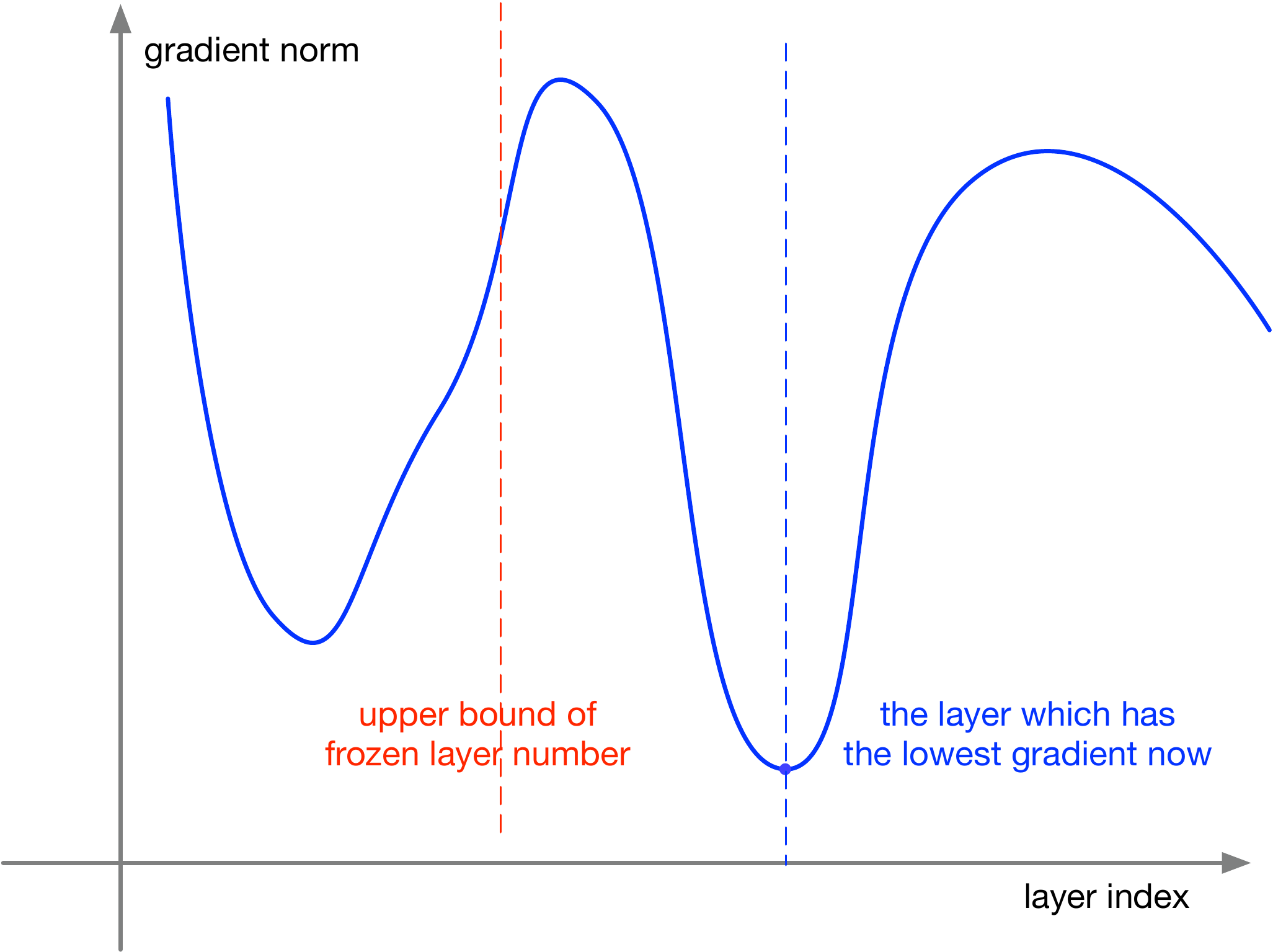}
    \caption{An example that the smallest gradient is not close to the input layer.}
    \label{fig:gradient_norm}
\end{figure}

\paragraph{Explanation of Equation \ref{eq:freeze}.} In numerical optimization, the weight with the smallest gradient norm converges first. With this assumption, we use the gradient norm as the indicator to identify which layers can be frozen on the fly. To verify this idea, we save the gradient norm for all layers at different iterations (i.e., epoch). With this analysis, we found that in the later phase of training, the pattern of gradient norm in different layers matches the assumption, but in the early phase, the pattern is random. Sometimes, we can even see that the gradient norm of those layers close to the output is the smallest. Figure \ref{fig:gradient_norm} shows such an example. If we freeze all layers preceding the blue dash line layer, the freezing is too aggressive since some layers have not converged yet. This motivates us further amend this naive gradient norm indicator. To avoid the randomness of gradient norm at the early phase of training, we use a tunable bound to limit the maximum number of frozen layers. We do not freeze all layers preceding the layer with the smallest gradient norm for the case in the figure. Instead, we freeze layers preceding the bound (the red color dash line).  

\paragraph{Deviation.} The term $L_{\text{frozen}}^{(T-1)} + \alpha(L - L_{\text{frozen}}^{(T-1)})$ in Equation \ref{eq:freeze}  can be written as:


\begin{align}
\label{eq:freeze_deviation}
\begin{split}
L_{\text{frozen}}^{(T)} = (1 - \alpha)^T[\frac{{\alpha}L}{1-\alpha} + \sum_{t=2}^{T}{\frac{{\alpha}L}{(1-\alpha)^t}}]\\
\end{split}
\end{align}

The deviation is as follows:
\begin{align}
&L_{\text{frozen}}^{(1)} = {\alpha}L\\
&L_{\text{frozen}}^{(2)} =(L - L_{\text{frozen}}^{(1)})\alpha + L_{\text{frozen}}^{(1)}\\ 
 &L_{\text{frozen}}^{(T)} =(L - L_{\text{frozen}}^{(T-1)})\alpha + L_{\text{frozen}}^{(T-1)}\\ 
&L_{\text{frozen}}^{(T)} ={\alpha}L + (1-\alpha)L_{\text{frozen}}^{(T-1)}\\
&\frac{L_{\text{frozen}}^{(T)}}{(1-\alpha)^T} = \frac{{\alpha}L}{(1-\alpha)^T} + \frac{L_{\text{frozen}}^{(T-1)}}{(1-\alpha)^{(T-1)}}\\
&\frac{L_{\text{frozen}}^{(T)}}{(1-\alpha)^T} = \frac{{\alpha}L}{(1-\alpha)} + \sum_{t=2}^{T}{\frac{{\alpha}L}{(1-\alpha)^{t}}}\\
\end{align}

\section{More Details of \code{AutoPipe}}
\label{appendix:load_balance}

\paragraph{Balanced Partition: Trade-off between Communication and Computational Cost.} Let us compute the communication cost in Figure \ref{fig:skip_connection}. The intermediate tensor from partition $k-2$ needs two cross-GPU communications to arrive to partition $k$. The parameter number of this intermediate tensor depends on the batch size and the Transformer model architecture. In $\text{BERT}_{\text{base}}$, the intermediate tensor width and height is the hidden feature size and sequence length, respectively (i.e., 1024, 512). If we use a batch size 300 in a pipeline, the total parameter number is $1024 \times 512 \times 300$. If we store it using \code{float32}, the memory cost is $0.63$ GB. The GPU-to-GPU communication bandwidth is $15.754$ GB (PCI 3.0, 16 lanes). Then one cross-GPU communication costs $40$ ms. In practice, the time cost will be higher than this value. Therefore, two cross-GPU communications cost around $100$ ms. To compare with the computation cost, we quantify the time cost for the forward propagation of a Transformer layer (12 million parameters), the time cost is around $35$ ms, meaning that the communication cost for skip connection is far more than a specific layer's computation cost. Compared to a slightly unbalanced partition in parameter number wise, $100$ ms is non-trivial. If we do not break the skip connection, the parameter number gap between different partitions is far less than 12 million (e.g., 4M or even less than 1 M). Therefore, this analysis explains partitioning without breaking the skip connection is a reasonable design choice. We also find that when the GPU device number in a machine is fixed (e.g., 8), the larger the model size is, the smaller the partition gap, which further indicates that our design's rationality.

\paragraph{Understanding Bubble in Pipeline.} In the main text, Figure~\ref{fig:bubble} depicts an example of running 4 micro-batches through a 4-device pipeline. 
Time flows from left to right, and each row denotes workload on one GPU device. \code{F} and \code{B} squares with the same color represent the forward and the backward pass time blocks of the same micro-batch. \code{U} represents the time block for updating parameters. Empty time blocks are bubbles. Assume that the load of the pipeline is evenly distributed amongst all devices. Consequently, all the time blocks during the forward pass are roughly in the same size, and similarly for backward time blocks. Note that the sizes of the forward time blocks can still differ from the backward ones. Based on these assumptions, we can estimate the per-iteration bubble size by simply counting the number of empty blocks during the forward and backward passes, respectively. In both the forward and backward pass, each device idles for $(K-1)$ time blocks. Therefore, the total bubble size is $(K-1)$ times per micro-batch forward and backward delay, which clearly decreases with fewer pipeline devices.

\paragraph{Relationship Between Number of Micro-batches per Mini-batch ($M$) and \texttt{DDP}.} To understand the reason why $M$ and \texttt{DDP} have mutual impacts, a thorough understanding of Section \ref{sec:hybrid} is needed first. In essence, \texttt{DDP} and pipelining has opposite requirement for $M$: \texttt{DDP} requires a relatively larger chunk of the bucket (smaller $M$) to overlap the communication (introduced in Section \ref{app:ddp}), while pipelining requires a larger $M$ to avoid bubble overhead (introduced in Section \ref{app:pipelining}). To further clarify, we must first remember that \texttt{DDP} must wait for the last micro-batch to finish its backward computation on a parameter before launching its gradient synchronization, then imagine two extreme cases. One case is that $M=1$, meaning the communication can be fully overlapped with computation using buckets. However, setting $M=1$ leads to a performance downgrade of pipelining (overhead of bubbles). Another extreme case is a very large $M$, then the communication time (labeled as green ``AR'' in Figure \ref{sec:hybrid}) may be higher than the computation time for a micro-batch (note that the width of a block in Figure \ref{sec:hybrid} represents the wall clock time). With these two extreme cases, we can see that there must be an optimal value of $M$ in a dynamical environment ($K$ and parameter number of active layers) of \autopipe, indicating that it is sub-optimal to fix $M$ during training. This explains the need for a dynamic $M$ for elastic pipelining.

\section{More details of \code{AutoDP}}
\label{app_sec:autodp}

\subsection{Data Redistributing}
\label{appendix:data_redistributing}

In standard data parallel-based distributed training, PyTorch uses \code{DistributedSampler} to make sure each worker in DP only load a subset of the original dataset that is exclusive to each other. The example code is as follows:

\code{self.train\_sampler = DistributedSampler(self.train\_dataset, \textcolor{red}{num\_replicas}=num\_replicas, \textcolor{red}{rank}=local\_rank)
}

Compared to this standard strategy, we made the following optimizations:

1. dynamic partition: the number of DP workers is increased when new pipelines have participated in DP. 
In order to guarantee that the data partition is evenly assigned after adding new pipes, the training dataset is repartitioned by rebuilding the \code{DistributedSampler} and setting new \code{num\_replicas} and \code{rank} as arguments. 

2. to reuse the computation of FP for frozen layers, we cached the hidden states in host memory and disk memory as well. 
Since the training requires to shuffle each epoch, the cache order of hidden features with respect to the order of original samples is different across different epochs.
In order to identify which data point a hidden feature belongs to, we build a sample unique ID by returning \code{index} in the \code{get\_item()} function of Dataset class.
With this unique ID, we can find a sample's hidden feature with O(1) time complexity during training.

3. when data is shuffled in each epoch, a data sample trained in the previous epoch may be moved to another machine for training in the current epoch. This makes the cache not reused across epochs. To address this issue, we fix a subset of entire samples in a machine and only do shuffle for this subset. This guarantees the shuffle during epochs is only executed inside a machine, thus the hidden feature's cache can be reused deterministically. To achieve this, rather than maintaining a global rank for DistributedSampler, we introduce \code{node\_rank} and \code{local\_rank}. \code{node\_rank} is used to identify which subset of samples a machine needs to hold. \code{local\_rank} is used by \code{DistributedSampler} to identify which part of the shuffle subset that a worker inside a machine should train.
Note that this does not hurt the algorithmic convergence property. Shuffling for multiple subsets obtains more randomness than randomness obtained by a global shuffle, which further increases the robustness of training. The only difference is that some parallel processes in distributed training are fixed in part of the shuffled datasets. If a training task does not need to shuffle the dataset across epochs, the above-mentioned optimization will not be activated.

\subsection{Skip Frozen Parameters in \code{AutoDP}}
\label{appendix:ddp}

To reduce communication cost, another method is to use PyTorch DDP API \footnote{See the internal API defined by PyTorch DDP: \url{ https://github.com/pytorch/pytorch/blob/master/torch/nn/parallel/distributed.py}, \texttt{\_set\_params\_and\_buffers\_to\_ignore\_for\_model()}.}. 
However, this API is temporally designed for Facebook-internal usage, and we must carefully calculate and synchronize the information regarding which parameters should be skipped, making our system unstable and difficult to be debugged. Our design avoids this issue and simplifies the system design.
Since \code{AutoPipe} stores $\mathcal{F}_{\text{frozen}}$ and $\mathcal{F}_{\text{pipe}}$ separately (introduced in Section \ref{sec:model_partition}), we can naturally skip the frozen parameters because \texttt{AutoDP} only needs to initialize the data parallel worker with $\mathcal{F}_{\text{pipe}}$.

\section{More Details of \code{AutoCache}}
\label{appendix:caching}

\begin{figure}[h!]
    \centering
    \includegraphics[width = 0.7 \linewidth]{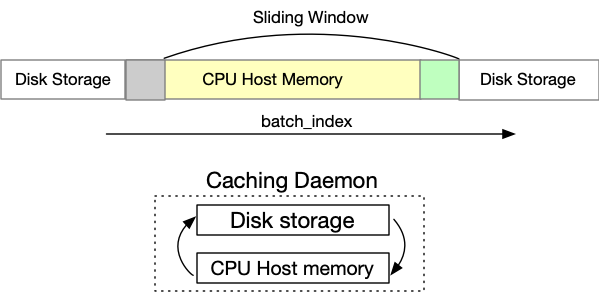}
    \caption{Hierarchical Caching}
    \label{fig:hierarchy_caching}
\end{figure}

\code{AutoCache} supports hierarchical caching. Figure \ref{fig:hierarchy_caching} shows our design. We maintain a sliding window to represent the maximum memory that the CPU host memory can hold, then move the window to prefetch the caching that the training requires and delete the caching that is consumed from the CPU host memory. In our implementation, we define the window size as the maximum batch number that the CPU host memory can hold. To avoid frequent memory exchange between disk storage and CPU host memory, we also define the block size that every time we prefetch (as the grey and green blocks are shown in the figure). In general, this hierarchical caching is useful when the training dataset is too large and exceeds the CPU host memory limit. However, we have to point out that this complex caching may not always be the optimal choice in the training system since the caching exchange itself may cost time. To this end, we suggest users of \autopipe\ using a relatively larger CPU host memory, which avoids activating the hierarchical caching and obtains faster training.

\section{More Experimental Results and Details}
\label{app_sec:experiments}

\subsection{Hyper-Parameters Used in Experiments}

\begin{table}[h!]
\footnotesize
\centering
\caption{Hyperparameters used in Experiments}
\label{table:hp_app}
\resizebox{\linewidth}{!}{
\begin{threeparttable}
\centering
 \begin{tabular}{lllc}
\toprule
Dataset & Model  &
Hyperparameters &
Comments\\
\midrule
\multirow{6}{*}{SQuAD} & \multirow{6}{*}{BERT}  &  batch size &  64 \\
&  & max sequence length & 512 \\
&   & learning rate &  \{1e-5, 2e-5, 3e-5, 4e-5, 5e-5\}\\
&   & epochs &  3 \\
&  & gradient accumulation steps &  1\\

\midrule
\multirow{7}{*}{ImageNet} & \multirow{7}{*}{ViT}  &  batch size &  400 \\
&  & image size & 224 \\
&   & learning rate & \{0.1, 0.3, 0.01, 0.03\}  \\
&   & weighs decay & 0.3 \\
&   & decay type &  cosine \\
&   & warmup steps &  2 \\
&   & epochs &  10 \\
\midrule
\multirow{7}{*}{CIFAR-100} & \multirow{7}{*}{ViT}  &  batch size &  320 \\
&  & image size & 224 \\
&   & learning rate & \{0.1, 0.3, 0.01, 0.03\}  \\
&   & weighs decay & 0.3 \\
&   & decay type &  cosine \\
&   & warmup steps &  2 \\
&   & epochs &  10 \\
\bottomrule 
\end{tabular}
\end{threeparttable}
    }
\end{table}
In Table \ref{table:hp_app}, we follow the same hyper-parameters used in the original ViT and BERT paper. Note that for ViT model, we use image size $224 \time 224$ for fine-tuning training.



\subsection{More Details of Speedup Breakdown}
\label{app_sec:speedup_breakdown}

\paragraph{Understanding the speed downgrade of \code{freeze only}.}

As shown in Figure~\ref{fig:breakdown}, the \code{Freeze Only} strategy is about 5\% slower than the \code{No Freeze} baseline. After the performance analysis, we found it is because \code{Freeze Only} changes memory usage pattern and introduced additional overhead in PyTorch's \code{CUDACachingAllocator} \footnote{To understand the design of this API, please refer to Section 5.3 in the original PyTorch paper \cite{paszke2019pytorch}. The source code is at \url{https://github.com/pytorch/pytorch/blob/master/c10/cuda/CUDACachingAllocator.h}}. More specifically, to reduce the number of expensive CUDA memory allocation operations, PyTorch maintains a \code{CUDACachingAllocator} that caches CUDA memory blocks to speed up future reuses. Without freezing, the memory usage pattern in every iteration stays consistent, and hence the cached memory blocks can be perfectly reused. After introducing layer freezing, although it helps to reduce memory footprint,  on the other hand, it might also change the memory usage pattern, forcing \code{CUDACachingAllocator} to split blocks or launch new memory allocations, which slightly slows down the training. In essence, this underlying mechanism of \texttt{PyTorch} is not tailored for freeze training. Customizing it for freeze training requires additional engineering efforts.

\subsection{Tuning $\alpha$ for ViT on ImageNet}

\begin{figure}[h!]
    \centering
    \includegraphics[width=\linewidth]{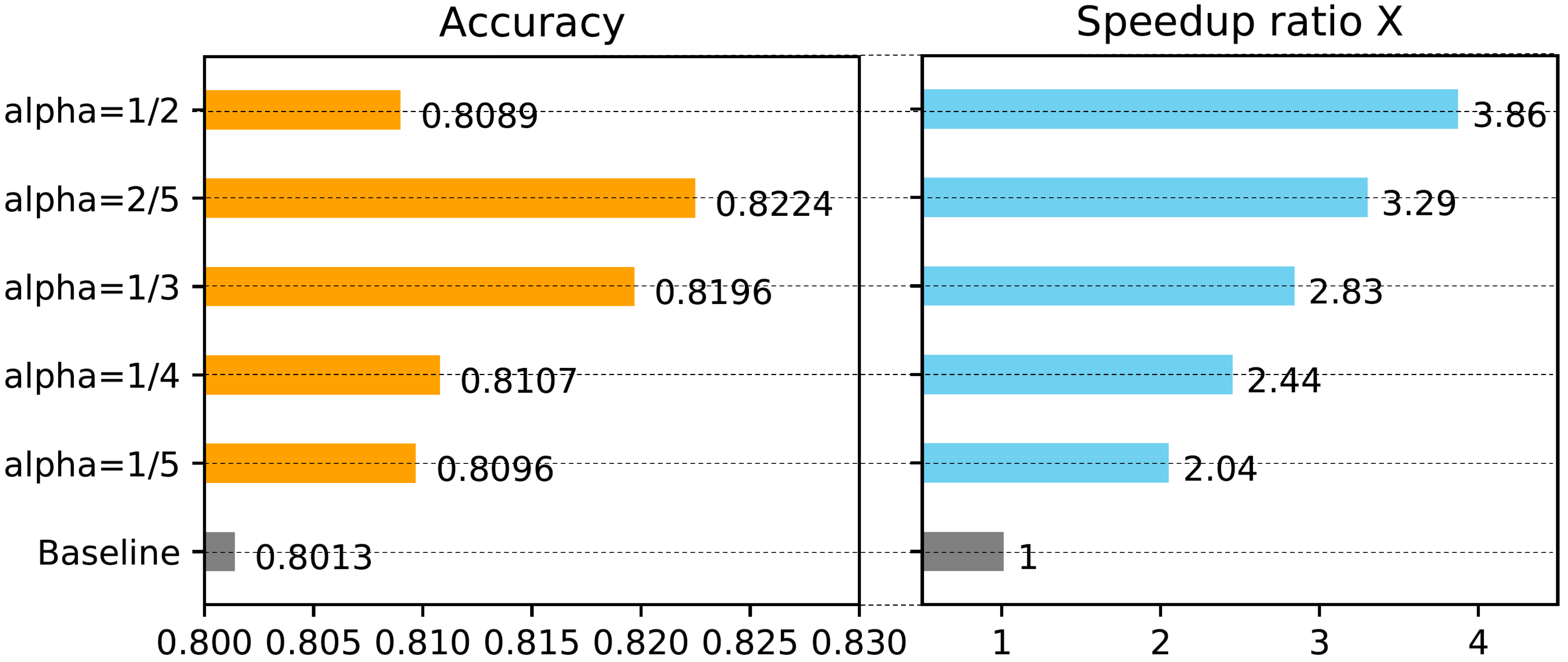}
    \setlength{\belowcaptionskip}{1cm}
    \caption{Tuning $\alpha$ for ViT on ImageNet}
\end{figure}

\subsection{The Method That Can Accurately Measure the Communication Cost}
\label{appendix:communication_cost}

Since PyTorch DDP overlaps communication with computation, the time difference between a local training iteration and a distributed training iteration does not faithfully represent the communication delay. Moreover, as DDP also organizes parameters into buckets and launches an \code{AllReduce} for each bucket, recording the start and finish time of overall communications is also insufficient. To correctly measure DDP communication delay, we combined the DDP communication hook with \code{CUDAFuture} callback. We developed a communication hook function that records a start \code{CUDA} event immediately before launching \code{AllReduce}. Then, in the \code{CUDAFuture} returned by the \code{AllReduce} function, we install a callback that records a finish \code{CUDA} event immediately after the non-blocking \code{CUDAFuture} completes. The difference between these two \code{CUDA} events represents the \code{AllReduce} communication delay of one bucket. We collected the events for all buckets and removed time gaps between buckets if there were any. The remaining duration in that time range accurately represents the overall DDP communication delay.

\subsection{Overheads of Pipe Transformation}
\label{sec:transformation_overhead}

\begin{table}[h!]
\caption{Overheads of pipe transformation (seconds)}
\label{table: overheads}
\begin{center}
\resizebox{\linewidth}{!}{
\begin{threeparttable}
\begin{tabular}{lcccc}
\toprule
\multirow{2}{*}{\textbf{Pipeline Transformation}} & \multirow{2}{*}{\textbf{Overall Time Cost}} & \multicolumn{3}{c}{\textbf{Dissect}}  \\
\cmidrule(lr){3-5}
 & & C & P & D\\
\midrule
initialization (length = 8) & 18.2 & 16.6 & 0.7 & 0.9 \\
length is compressed from 8 to 4 & 10.2 & 8.3 & 1.3 & 0.6 \\
length is compressed from 4 to 2 & 5.5 & 3.8 & 2.1 & 0.7 \\
length is compressed from 2 to 1 & 9.5 & 2.3 & 6.1 & 1.0 \\
\bottomrule
\end{tabular}
\end{threeparttable}}
\end{center}
\begin{tablenotes}
      \footnotesize
      \item *C - creating CUDA context; P - Pipeline Warmup; D - DDP.
\end{tablenotes}
\label{table:transformation_overhead}
\end{table}
We have verified the time cost of pipeline transformation. The result in Table \ref{table:transformation_overhead} shows that the overall cost of pipeline transformation is very small (less than 1 minute), compared to the overall training time. Therefore, we do not consider further optimization. 

\section{Discussion}
\label{appsec:discussion}

\textbf{Pretraining v.s. Fine-tuning}: Given that the model architectures in pre-training and fine-tuning are the same, we do not need to change the system design. Running larger Transformers (over 32 layers) is straightforward because almost all giant Transformer models are designed by simply stacking more transformer encoder layers. \autopipe\ can serve as a training system for both pre-training and fine-tuning training. We plan to run our training system on more models and datasets in both settings.

\textbf{Designing better freeze algorithm}: Our proposed algorithm is simple, yet it proves to be effective on various tasks. However, we believe that further developments to the freeze algorithm may lead to better generalization and obtain higher accuracy. 

\textbf{versatility}: \autopipe\ training system can also be used on other algorithms that gradually fix portions of neural networks. For example, cross-silo federated learning, layer-by-layer neural architecture search, and pruning large DNNs are all potential use cases of our training system. We will explore the training acceleration for these scenarios in our future works.

\end{document}